\documentclass{article}

\PassOptionsToPackage{numbers, sort}{natbib}

\usepackage[table]{xcolor}
\usepackage[preprint]{neurips_2025}

\usepackage[utf8]{inputenc} %
\usepackage[T1]{fontenc}    %
\usepackage{hyperref}       %
\usepackage{url}            %
\usepackage{booktabs}       %
\usepackage{amsfonts}       %
\usepackage{nicefrac}       %
\usepackage{microtype}      %

\usepackage{amsmath,amsfonts,bm}

\def\eqref#1{equation~\ref{#1}}

\def\1{\bm{1}}

\DeclareMathAlphabet{\mathsfit}{\encodingdefault}{\sfdefault}{m}{sl}
\SetMathAlphabet{\mathsfit}{bold}{\encodingdefault}{\sfdefault}{bx}{n}

\def\sD{{\mathbb{D}}}

\def\sS{{\mathbb{S}}}

\usepackage{graphicx}
\usepackage{subfig}
\usepackage{cleveref}
\usepackage{todonotes}

\usepackage{multirow}

\usepackage{array} %
\usepackage{makecell}
\usepackage{adjustbox} %
\usepackage{tablefootnote} %
\usepackage{threeparttable}
\usepackage{wrapfig}

\title{Effective Data Pruning through Score Extrapolation}
\author{Sebastian Schmidt\textsuperscript{1, 3,\textdagger} \quad Prasanga Dhungel\textsuperscript{1,\textdagger} \quad Christoffer Löffler\textsuperscript{4} \quad Björn Nieth\textsuperscript{6} \\ \textbf{Stephan Günnemann}\textsuperscript{\textbf{1,2,5}} \quad \textbf{Leo Schwinn}\textsuperscript{\textbf{1,2}}\\
$^1$ Technical University of Munich, School of Computation, Information and Technology \\
$^2$ Munich Data Science Institute \quad
 $^3$ BMW Group \\ $^4$ Pontificia Universidad Católica de Valparaíso \quad $^5$ Pruna AI 
 \\
 $^6$ Friedrich-Alexander Universität Erlangen-Nürnberg
}

\begin{document}

\maketitle

\renewcommand*{\thefootnote}{\textdagger}
\footnotetext{These authors contributed equally to this work.}
\renewcommand*{\thefootnote}{}
\footnotetext{Corresponding authors {\tt sebastian95.schmidt@tum.de, prasanga.dhungel@tum.de}}

\renewcommand*{\thefootnote}{\arabic{footnote}}
\setcounter{footnote}{0}

\newcommand{\ours}{\textsc{Extra\,}}
\begin{abstract}

Training advanced machine learning models demands massive datasets, resulting in prohibitive computational costs. To address this challenge, data pruning techniques identify and remove redundant training samples while preserving model performance. Yet, existing pruning techniques predominantly require a full initial training pass to identify removable samples, negating any efficiency benefits for single training runs. 
To overcome this limitation, we introduce a novel importance score extrapolation framework that requires training on only a small subset of data. We present two initial approaches in this framework\footnote{\url{https://github.com/prasangadhungel/Data-Pruning-with-Extrapolated-Scores}}—k-nearest neighbors and graph neural networks—to accurately predict sample importance for the entire dataset using patterns learned from this minimal subset. We demonstrate the effectiveness of our approach for $2$ state-of-the-art pruning methods (Dynamic Uncertainty and TDDS), $4$ different datasets (CIFAR-10, CIFAR-100, Places-365, and ImageNet), and $3$ training paradigms (supervised, unsupervised, and adversarial). 
Our results indicate that score extrapolation is a promising direction to scale expensive score calculation methods, such as pruning, data attribution, or other tasks.

\end{abstract}

\section{Introduction}

In recent years, the demand for large and comprehensive datasets has grown rapidly. This is particularly evident in the development of advanced models, such as large language models (LLMs)~\cite{Minaee2024}, and other forms of foundation models~\cite{Kirillov2023, gotz2025efficient, gotz2025byte}, which require vast amounts of data to train.

In this context, dataset pruning has emerged as a valuable technique to optimize the training process and improve the efficiency of model development. By scoring individual data points by their importance and only selecting the most informative, dataset pruning aims to improve training efficiency while maintaining model performance. This approach is particularly beneficial in scenarios where the available dataset is vast, and the computational resources required for training the model on the entire dataset are significant, such as autonomous driving~\cite{Schmidt2020, schmidt2024generalized}.

However, existing pruning approaches generally require training a model on the full dataset to calculate data importance scores~\cite{he2024large,zhang2024spanning}. As full, large-scale training runs are often conducted only once for a specific model, the effort required for this initial full-dataset training will often exceed the benefits gained from the pruning process, rendering the approach impractical. 

To address this limitation, we propose a framework for score extrapolation. Our framework limits the expensive score computation to a small initial subset of the full training data. Subsequently, we extrapolate scores from the small subset to the entire dataset. By \emph{avoiding the need for a full-dataset training}, score extrapolation substantially accelerates the dataset pruning and enables it for practical applications. Beyond immediate pruning applications, our findings suggest that importance score extrapolation offers a scalable approach that may be applied to other techniques involving expensive per-sample evaluations, including data attribution methods~\cite{ilyas2022datamodels} or influence estimation~\cite{koh2017understanding}.

Our contribution can be summarized as follows:
\begin{itemize}
    \item We provide a novel framework for score extrapolation, which can be used to extrapolate expensive score extrapolation methods to unseen data samples, \emph{significantly reducing computational effort}.
    \item In the scope of this framework, we introduce two extrapolation techniques based on k-nearest-neighbors (KNN) and graph neural networks (GNN).
    \item In an extensive empirical study, we showcase the effectiveness of our extrapolation-based approaches for \textbf{2} state-of-the-art pruning methods, i.e., Dynamic Uncertainty (DU)~\cite{he2024large}, Temporal Dual-Depth Scoring (TDDS)~\cite{zhang2024spanning}, \textbf{4} datasets (CIFAR-10~\cite{cifar}, CIFAR-100~\cite{cifar}, Places-365~\cite{zhou2017places}, ImageNet~\cite{Deng2009}), and \textbf{3} different training paradigms (supervised, unsupervised, adversarial).
\end{itemize}

\section{Related Work}
Data pruning is usually costly to apply, as it requires either a full training or costly optimizations. We review the literature on data pruning in supervised and adversarial training.

\textbf{Data Pruning.}
Data pruning, or coreset selection \cite{sorscher2022beyond,he2024large,Zhang2024,sener2017active,feldman2020neural,Guo2022}, aims to keep a small, representative subset of training data that preserves model performance while reducing computational costs, which is specifically important when training is costly~\cite{gao2023generalizing, gaoneural, nieth2024large}. Several efficient strategies have emerged, which can be grouped into three categories \cite{tan2023data} and either estimate importance or difficulty scores, use geometric calculation, or employ optimization. Importantly, approaches usually require full training on the dataset \cite{Zhang2024,he2024large,toneva2018empirical} to estimate importance scores or to create a latent space for pruning.

\textbf{Pruning Based on Importance.}
Methods from this category assign scores to samples based on their utility for training, typically retaining the highest-ranked examples. 
Common techniques include GradNd \cite{paul2021deep} or TDDS \cite{zhang2024spanning} and rely on gradients, while others, like EL2N \cite{paul2021deep}, use prediction errors of the model instead.
Coleman et al.~\cite{coleman2019selection} use the entropy from proxy models for ranking, while forgetting scores track transitions between correct and incorrect classifications, suggesting that frequently forgotten samples are informative. AUM \cite{pleiss2020identifying} identifies mislabeled measures by the average margin between the true class logit and the highest logit. DU \cite{he2024large} assesses uncertainty through the standard deviation of predictions during training. \\
An alternative approach measures importance by the impact of sample removal. The memorization score evaluates changes in confidence for the true label with and without the sample. The MoSo~\cite{tan2023data} score calculates the change in empirical risk, while \cite{feldman2020neural, yang2022dataset} use influence functions \cite{koh2017understanding} to gauge the influence of a sample on generalization performance. 
While this category of pruning approaches is usually the most effective, it comes with a costly full dataset training to estimate the pruning scores.

\textbf{Pruning Based on Geometry.} 
Other approaches utilize geometric properties or the distribution of the data. Herding \cite{Welling2009,Chen2010} and Moderate \cite{xia2022moderate} calculate distances in feature space. Some approaches \cite{Huang2024,Mirzasoleiman2020} combine gradients with distance calculations.

For instance, \cite{har2006maximum} constructs a coreset by approximating the maximum margin separating hyperplane. Self-Supervised Pruning~\cite{sorscher2022beyond} applies $k$-means clustering in the embedding space of a self-supervised model and ranks samples by their cosine similarity to cluster centroids, retaining samples with the lowest similarity. 
The $k$-center method~\cite{sener2017active} selects centers that minimize the maximum distance between any sample and its nearest center. Coverage-based approaches~\cite{zheng2022coverage, griffin2024zero} aim to maximize sample diversity while minimizing redundancy.

\textbf{Pruning Based on Optimization.} 
Some methods formulate pruning as an optimization problem. \citet{borsos2020coresets} used greedy forward selection to solve a cardinality-constrained bilevel optimization problem for subset selection. GLISTER~\cite{killamsetty2021glister} performs coreset selection via a greedy-Taylor approximation of a bilevel objective while simultaneously updating model parameters. GradMatch~\cite{killamsetty2021grad} defines a gradient error term and minimizes it using the Orthogonal Matching Pursuit algorithm.

\textbf{Adverserial Robustness.} 
Adversarial training~\cite{madry2017towards, bungert2021clip, wang2023better, altstidl2024raising, xhonneux2024efficient} and other robustification methods~\cite{schwinn2021identifying, schwinn2022improving, Schuchardt2021_Collective, Schuchardt2022_Invariance}, generally involve expensive optimization~\cite{madry2017towards, schwinn2021dynamically, schwinn2023exploring, kollovieh2023assessing, Schuchardt2023_Localized}, and thus entail large computational overhead. Yet, data pruning in adversarial training remains relatively underexplored. Existing approaches can be grouped into three broad categories. The first focuses on dynamically selecting which samples to apply adversarial perturbations to during training \cite{hua_bullettrain_2021, chen_data_2024}. The second incorporates coreset-based pruning methods to reduce training data while preserving robustness \cite{killamsetty_grad-match_2021, dolatabadi_adversarial_2023}. The third relies on heuristic pruning strategies to discard samples deemed less useful for robust learning \cite{kaufmann_efficient_2022, li_less_2023}.

\textbf{Orthogonal Works.} 
Beyond data pruning, identifying important subsets of data is a fundamental problem in various machine learning paradigms, including continual learning \cite{Wang2023}, data distillation \cite{Wang2018,Holder2021}, and active learning \cite{kirsch2022unifying, schmidt2024unified, schmidt2025joint, loeffler2022iale, loeffler2023active}. While active learning focuses on selecting samples to label, data pruning addresses the challenge of selecting which samples to retain for training. 

While data pruning methods lead to effective data reduction, approaches usually require training on the full dataset or complex optimization to estimate which samples to prune. Only a few works consider the time-of-accuracy problem to increase time efficiency, including RS2 \cite{Okanovic2024}, which proposes repeated random sampling, \cite{Tang2023} selects data online during training by gradient evaluations or uses proxy models~\cite{coleman2019selection}. Here, we extend our previous short paper~\cite{nieth2024large} and address the prohibitively expensive training on the full datasets is an unaddressed problem, which we address in this work. 

\section{Score Extrapolation}
\begin{figure}[t!]
    \vskip -0.2cm
    \centering
    \includegraphics[width=1\linewidth]{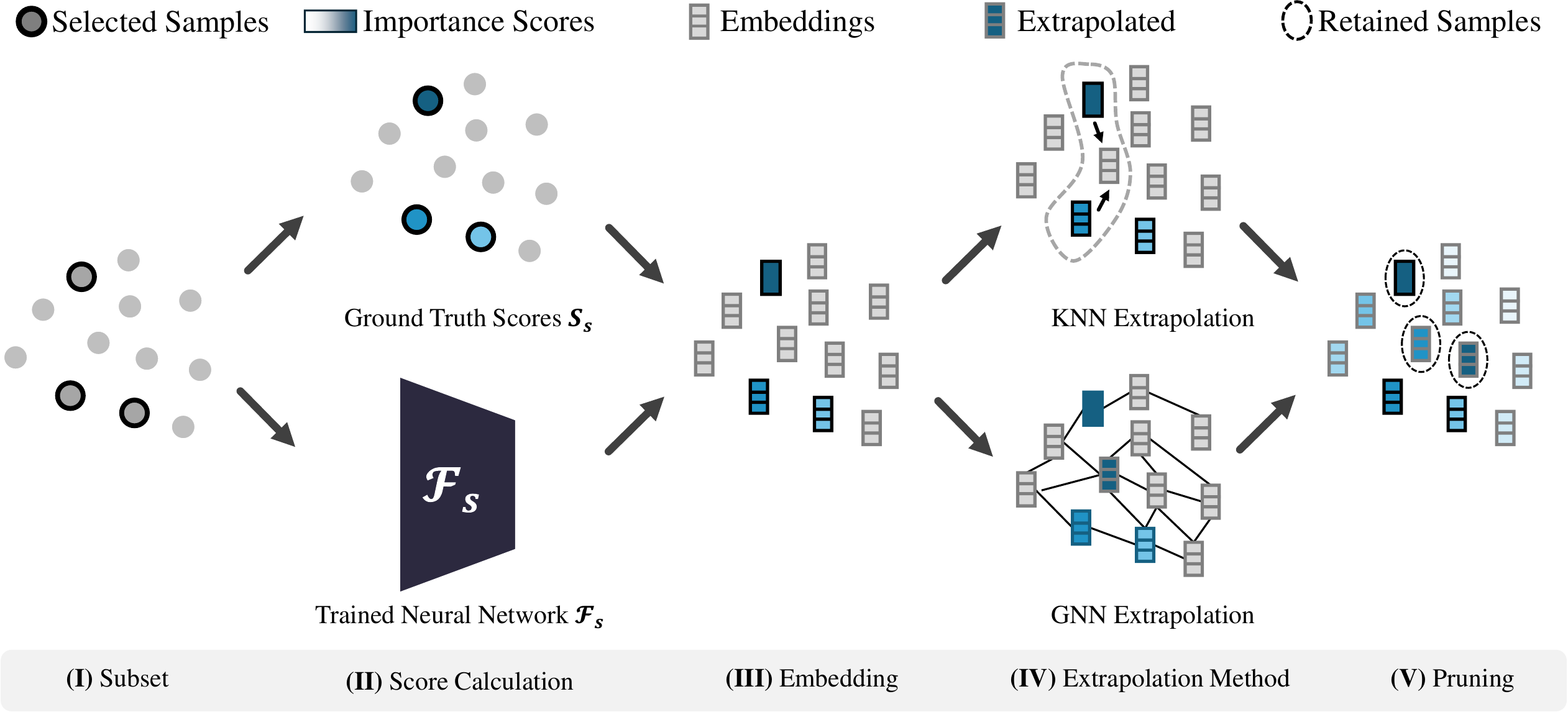}
    \caption{Extrapolation concept overview. \textbf{(I)} We start by randomly selecting a subset $\sD_s$ of $m$ samples out of the full dataset $\sD$ compromising $n$ samples (where $m << n$). \textbf{(II)} We train a neural network $F_s$ on the selected subset and calculate ground truth importance scores $S_S$ with the selected pruning method during the training run. \textbf{(III)} Using the trained model, we map the whole dataset $\sD$ to the embedding space of the network. \textbf{(IV)} Subsequently, we select an extrapolation method and train it to extrapolate scores on $\sD_s$ and the calculated ground truth scores $S_s$. \textbf{(V)} Finally, we use the original and extrapolated scores to perform the pruning task, selecting the top-k samples exhibiting the highest importance scores.}
    \label{fig:concept}
    \vskip -0.2cm
\end{figure}

Existing methods \cite{Zhang2024,he2024large,paul2021deep} in the area of data pruning require complete training on the full dataset or rely on costly optimizations \cite{killamsetty2021glister}, which makes them prohibitively expensive.

To address the computational challenges of estimating data importance scores for large-scale data sets, we propose a score extrapolation framework.
Within this framework, we compute the importance scores on \emph{only} on a minor subset of the large-scale dataset and extrapolate scores for the remaining data points efficiently. The concept is shown in \Cref{fig:concept} and comprises the following steps: At first, a subset of the large-scale dataset is selected \textbf{(I)}. Based on this subset, a model is trained to calculate importance scores based on a chosen extrapolation method \textbf{(II)}. To extrapolate the scores of the remaining samples, we generate the embeddings of the dataset based on the trained model \textbf{(III)}. With the calculated scores of the subset and the embeddings, we use one of the proposed extrapolation methods based on either KNN or a GNN \textbf{(IV)} to estimate scores for the remaining samples. In the last step \textbf{(V)}, we perform the regular pruning task with our extrapolated scores.

\textbf{Subset Definition.}
Rather than directly calculating the importance score for each sample in the dataset, we start by selecting a training subset $\sD_s \subset \sD$ of cardinality $m$ from the full dataset $\sD$, comprising $n$ samples, where $m << n$. Our goal is to extrapolate data importance scores for the residual set $\sD_r = \sD \setminus \sD_s$ containing the remaining $n - m$. We do not assume specific knowledge about our dataset $\sD$ and therefore use random sampling for generating $\sD_s$. 

\textbf{Score Calculation and Model Training.}
Next, we select a suitable data pruning method from which to extrapolate scores. The choice of pruning method is critical, as the quality of extrapolated scores is inherently upper-bounded by the chosen method. We then train a model $\mathcal{F}_s$ on the training subset $\sD_s$, using the same setup as the pruning method would normally use for the full dataset. Through the training process, we obtain ``ground truth'' data importance scores $S_s \in \mathbb{R}^m$ for the subset $\sD_s$.

\textbf{Score Extrapolation.}
Based on the initial computed scores $S_s$, we apply an extrapolation schema to generate the scores $S_r\in \mathbb{R}^{n-m}$ for the residual set $\sD_r$. In this work, we propose two extrapolation schemes: a basic KNN approach and a more advanced interpolation based on GNNs. Both concepts are illustrated in \Cref{fig:concept}. Finally, all samples in \(\sD\) are assigned importance scores: samples in \(\sD_s\) retain their directly computed values, while scores for \(\sD_r\) are inferred. Once scores are computed for the full dataset, they can be used in the respective downstream tasks. In this work, we specifically focus on data pruning. However, the general framework could also be applied to other settings, such as data attribution~\cite{ilyas2022datamodels}, influence functions~\cite{koh2017understanding}.

This strategy significantly reduces the computational cost of scoring large datasets. Instead of training on the full dataset, scores are computed only on a small subset, while the remaining samples are efficiently approximated through extrapolation.

\subsection{Extrapolation with KNN}
\label{section: knn}
 
In the following, we propose two methods to extrapolate scores. First, as our baseline, we propose a simple KNN-based approach to estimate the importance scores.  
We start from the subset $\sD_s$ and its associated importance scores $S_s$. Next, we utilize the encoder of our trained model $\mathcal{F}_s: \mathbb{R}^d \to \mathbb{R}^{d'}$ to transform $d$ dimensional input samples $x \in \mathbb{R}^d$ into the $d'$-dimensional embedding space of the encoder.
Then, we compute the score for each data point $x \in \sD_r$ as the average of the scores of its $k$ nearest neighbors in the embedding space. Finally, the extrapolated importance score $S_{knn}$ for a sample $x$ can be expressed as
\begin{equation}
S_{knn}(x) = \frac{\sum_{i=1}^{k} \exp(-D(\mathcal{F}_s(x), \mathcal{F}_s(x_{\pi_i(x)}))) S_{\pi_i(x)}}{\sum_{i=1}^{k} \exp(-D(\mathcal{F}_s(x), \mathcal{F}_s(x_{\pi_i(x)})))},
\label{eq:extrapolation}
\end{equation}
where $\pi_i(x)$ represents the index of the $i$-th nearest neighbor of $x$, and $D(\cdot, \cdot)$ denotes a chosen distance metric (i.e., Euclidean distance). This weighted average, based on the structure of the embedding space, ensures that the extrapolated scores maintain the local structure of the dataset.

\subsection{Extrapolation with GNN}\label{section:gnn-extrapolation}
While KNN-based extrapolation serves as a simple approach to extrapolate information from the sampled subset $\sD_s$ to the residual data $\sD_r$, it lacks the ability to model complex interactions among data points.
To address this limitation, we additionally propose a more powerful extrapolation method based on GNNs that can capture higher-order relationships in the dataset through message passing. We construct an undirected graph $\mathcal{G} = (\mathcal{V}, \mathcal{E})$, where each sample in $\sD$ represents a vertex in $\mathcal{V}$, and edges $\mathcal{E}$ are formed between each sample and its $k$ nearest neighbors in the embedding space. 
The node embeddings are defined based on the sample embeddings of the model $\mathcal{F}_s$ and are combined with the one-hot encoded class labels for supervised tasks.

In addition, we define edge weights based on the latent space distance to its neighbors encoded as $\exp(-d(\mathcal{F}_s(x), \mathcal{F}_s(x_{\pi_i(x)})))$. The adjacency matrix $\mathcal{A}$ is constructed based on these edges. 

To learn the interactions between the data samples, we employ a simple GNN $\mathcal{F}_{\mathcal{G}}(\mathcal{A}, \mathcal{V}; \theta)$ with weights $\theta$ that consists of three layers of Graph Convolutional Networks (GCNs) as described by \cite{kipf2016semi} and directly predicts the importance score of the sample nodes given our defined node embeddings $\mathcal{V}$ and adjacency matrix $\mathcal{A}$.

To scale training in large graphs, we employ neighbor sampling \cite{hamilton2017inductive} to generate mini-batches of nodes and their local neighborhoods during training. 

Importantly, GNNs are significantly less computationally expensive than the task model, which we analyze in our experiments.
The GNN outputs a vector of predicted scores $\mathbb{R}^n$, i.e, a scalar score for each node. Since we only have the reference score $S_s(x)$ for samples in the training dataset $\sD_s$, we compute the mean square loss only over these nodes as 

\begin{equation} \mathcal{L} = \frac{1}{|\sD_s|} \sum_{x_i \in \sD_s} \left( \mathcal{F}_{\mathcal{G}}(\mathcal{A}, \mathcal{V}; \theta)_i - S(x_i) \right)^2, \end{equation}

where $\mathcal{F}_{\mathcal{G}}(\mathcal{A}, \mathcal{V}; \theta)_i$ is the prediction for index $i$.
After training, we use the GNN to infer scores for all samples in $\sD \setminus \sD_s$. The message passing in the GNN allows the model to leverage the structural information of the entire dataset, potentially leading to more accurate extrapolation of the scores. 

\section{Evaluation}
\label{sec:Experiments}
In the following section, we aim to evaluate our importance score extrapolation for different scores, tasks, and datasets. The primary objective of this paper is to evaluate the practical feasibility of score extrapolation. We present this as foundational research that characterizes the approach's strengths and limitations, upon which future optimization efforts can build. Our experiments are designed to analyze the core properties of the method rather than to maximize performance metrics. We have three main goals for score extrapolation: 1) \textit{reduce} the computation time compared to standard pruning, 2) \textit{maintain} downstream task performance, and 3) show high \textit{correlation} to the original scores which are generated by training a model and estimating the scores on the full set $\sD$.
To examine these properties, we evaluate the tasks of classic data pruning for labeled data, unsupervised data pruning, and adversarial training. 

\textbf{Scores.}
In our experiments, we extrapolate two state-of-the-art pruning methods, DU \cite{he2024large} and TDDS \cite{zhang2024spanning}. Both methods require training on the full dataset for several epochs to obtain reliable scores. They also store logits per sample at each epoch; DU retains only the softmax logits for the correct class, whereas TDDS approximates gradients from full logit outputs, making it more memory-intensive. Details on these scores are provided in \Cref{ap:Scores}.

\textbf{Supervised Data Pruning.}
We first evaluate score extrapolation on supervised data pruning. The objective of this task is to minimize the amount of training data while maintaining model performance as much as possible. Since we focus on large datasets, we use synthetic CIFAR-100 1M \cite{cifar,wang2023better}, Places 365 \cite{zhou2017places}, and ImageNet 1k \cite{Deng2009}. 
We compare our extrapolated scores with the original pruning approaches and random pruning and examine the different initial subset sizes $\sD_s$ ranging from  10\%, to 25\%. If not stated otherwise, we use $20\%$ of the full dataset for $\sD_s$ in all experiments. More details are given in \Cref{ap:experimentSetup}. All experiments were conducted with $3$ different random initializations.

 \begin{table}[bh]
\centering
\vskip -0.3cm
\small
\setlength{\tabcolsep}{2pt}
\renewcommand{\arraystretch}{1.1}  %
\caption{Accuracy (± std.\ error) and inference time (± std.\ error, in minutes) for various datasets, at the highest pruning percentages, where the pruning algorithms still outperform random pruning. Small subset corresponds to 10\% of the initial dataset, while medium corresponds to 20\% for Synthetic CIFAR and ImageNet and 25 \% for Places365.}
\vskip 0.2cm
\label{tab:pruning-results}
\resizebox{1\columnwidth}{!}{
\begin{tabular}{*{9}{c}}
\toprule
\makecell{\bfseries Dataset} 
 & \makecell{\bfseries Prune\\\bfseries \%} 
 & \makecell{\bfseries Method}
 & \makecell{\bfseries Original}
 & \makecell{\bfseries GNN\\\bfseries Medium Set}
 & \makecell{\bfseries KNN\\\bfseries Medium Set}
 & \makecell{\bfseries GNN\\\bfseries Small Set}
 & \makecell{\bfseries KNN\\\bfseries Small Set}
 & \makecell{\bfseries Random} \\
\midrule
\shortstack{Imagenet}
  & \shortstack{50}
  & \shortstack{DU}
  & \shortstack{59.08 ± 0.07\\\textcolor{gray}{1367 ± 19 }}
  & \shortstack{58.89 ± 0.16\\\textcolor{gray}{829 ± 15 }}
  & \shortstack{58.74 ± 0.14\\\textcolor{gray}{687 ± 16 }}
  & \shortstack{58.30 ± 0.17\\\textcolor{gray}{701 ± 12 }}
  & \shortstack{58.51 ± 0.14\\\textcolor{gray}{581 ± 10 }}
  & \shortstack{58.56±0.06\\\textcolor{gray}{-}} \\
\addlinespace
\shortstack{Places365}
  & \shortstack{50}
  & \shortstack{DU}
  & \shortstack{42.85 ± 0.09\\\textcolor{gray}{2416 ± 26 }}
  & \shortstack{42.51 ± 0.14\\\textcolor{gray}{1631 ± 18 }}
  & \shortstack{42.30 ± 0.12\\\textcolor{gray}{1330 ± 14 }}
  & \shortstack{42.32 ± 0.13\\\textcolor{gray}{1394 ± 15 }}
  & \shortstack{42.36 ± 0.11\\\textcolor{gray}{1075 ± 13 }}
  & \shortstack{42.14 ± 0.09\\\textcolor{gray}{-}} \\
\addlinespace
\shortstack{Places365}
  & \shortstack{95}
  & \shortstack{TDDS}
  & \shortstack{34.61 ± 0.17\\\textcolor{gray}{1741 ± 22 }}
  & \shortstack{33.96 ± 0.26\\\textcolor{gray}{917 ± 17 }}
  & \shortstack{33.56 ± 0.19\\\textcolor{gray}{621 ± 13 }}
  & \shortstack{33.58 ± 0.21\\\textcolor{gray}{668 ± 11 }}
  & \shortstack{33.29 ± 0.18\\\textcolor{gray}{353 ± 8 }}
  & \shortstack{32.89 ± 0.13\\\textcolor{gray}{-}} \\
\addlinespace
\shortstack{Synthetic\\CIFAR-100}
  & \shortstack{95}
  & \shortstack{DU}
  & \shortstack{71.42 ± 0.21\\\textcolor{gray}{412 ± 9 }}
  & \shortstack{69.62 ± 0.26\\\textcolor{gray}{374 ± 6 }}
  & \shortstack{69.13 ± 0.22\\\textcolor{gray}{158 ± 5 }}
  & \shortstack{68.60 ± 0.24\\\textcolor{gray}{333 ± 6 }}
  & \shortstack{68.81 ± 0.23\\\textcolor{gray}{114 ± 5 }}
  & \shortstack{67.38 ± 0.19\\\textcolor{gray}{-}} \\
\addlinespace
\shortstack{Synthetic\\CIFAR-100}
  & \shortstack{95}
  & \shortstack{TDDS}
  & \shortstack{72.51 ± 0.18\\\textcolor{gray}{441 ± 10 }}
  & \shortstack{71.16 ± 0.28\\\textcolor{gray}{387 ± 6 }}
  & \shortstack{70.80 ± 0.24\\\textcolor{gray}{168 ± 6 }}
  & \shortstack{69.24 ± 0.22\\\textcolor{gray}{343 ± 7 }}
  & \shortstack{69.10 ± 0.21\\\textcolor{gray}{122 ± 5 }}
  & \shortstack{67.38 ± 0.20\\\textcolor{gray}{-}} \\
\bottomrule
\end{tabular}
}
\end{table}

\Cref{tab:pruning-results} illustrates the effect of our extrapolation schemes across the different datasets and pruning approaches. 
As expected, the results indicate that the extrapolated scores perform slightly worse than the original scores. However, score extrapolation is considerably faster in our experiments, while requiring less data for initial model training. The time measurement contains \emph{all} steps, including the training of the model for initial scores $S_s$, possible extrapolation approaches, and the training of the final model on the pruned subset. 

Notably, the medium-sized training subsets $\sD_s$ demonstrate strong performance, while the small subset already shows decent results and generally outperforms random pruning. GNN-based extrapolation generally performs better than using KNN. Still, KNN extrapolation is \emph{pareto optimal} w.r.t. time-accuracy trade-offs (see Time Optimality). 

\begin{figure}[bt]
\centering
\subfloat[Places 365\label{fig:pruningPlaces}]{%
   \includegraphics[width=0.49\textwidth]{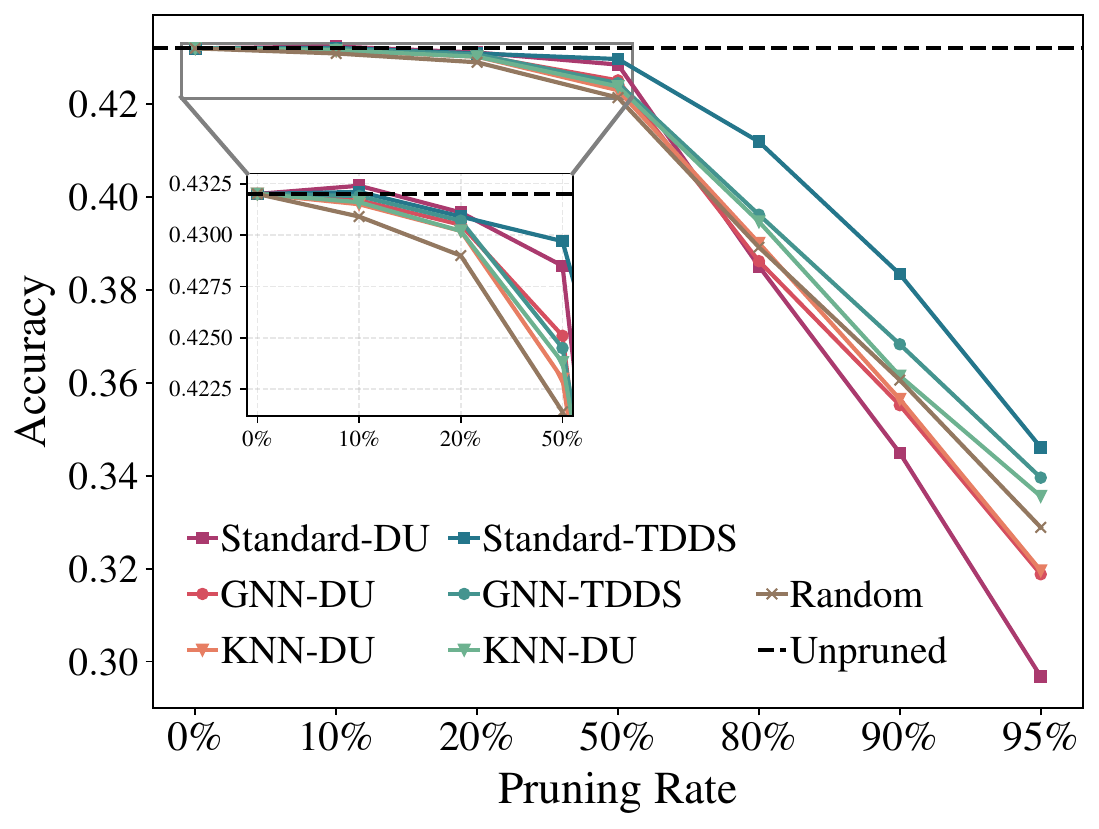}
}%
\subfloat[ImageNet\label{fig:pruningImageNet}]{%
 \includegraphics[width=0.49\textwidth]{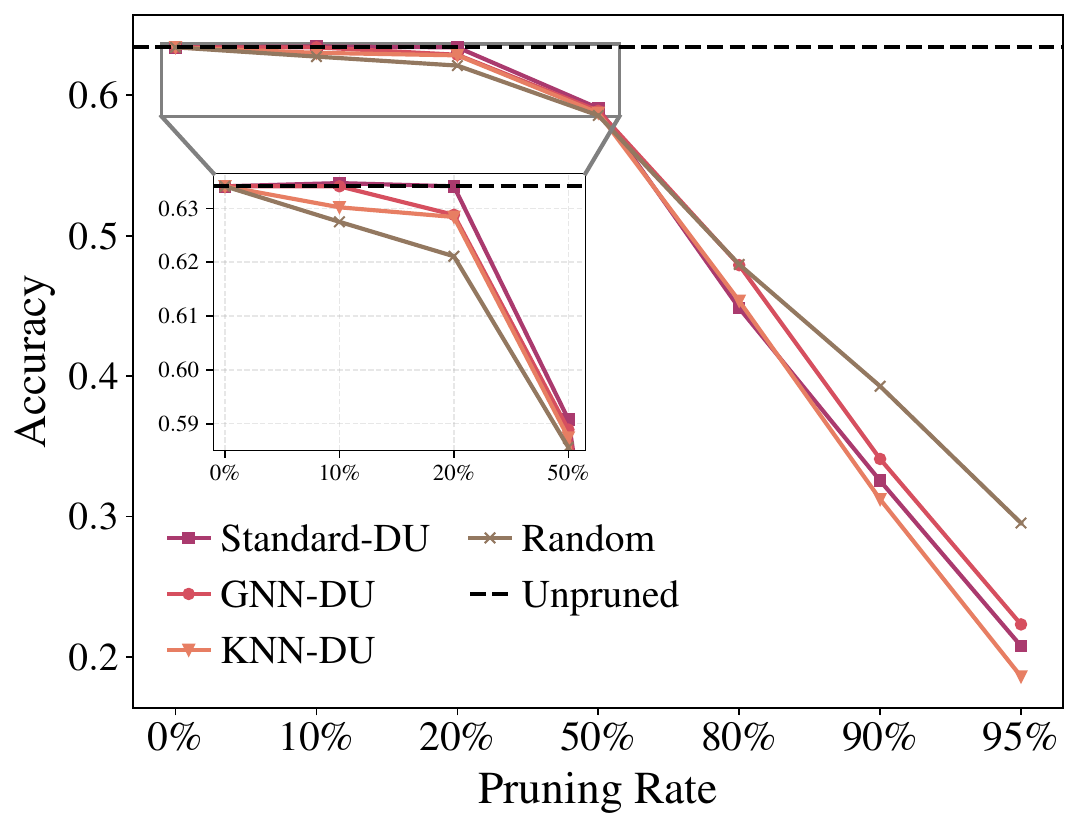}
}

\subfloat[Synthic CIFAR-100\label{fig:pruningCifar100}]{%
   \includegraphics[width=0.49\textwidth]{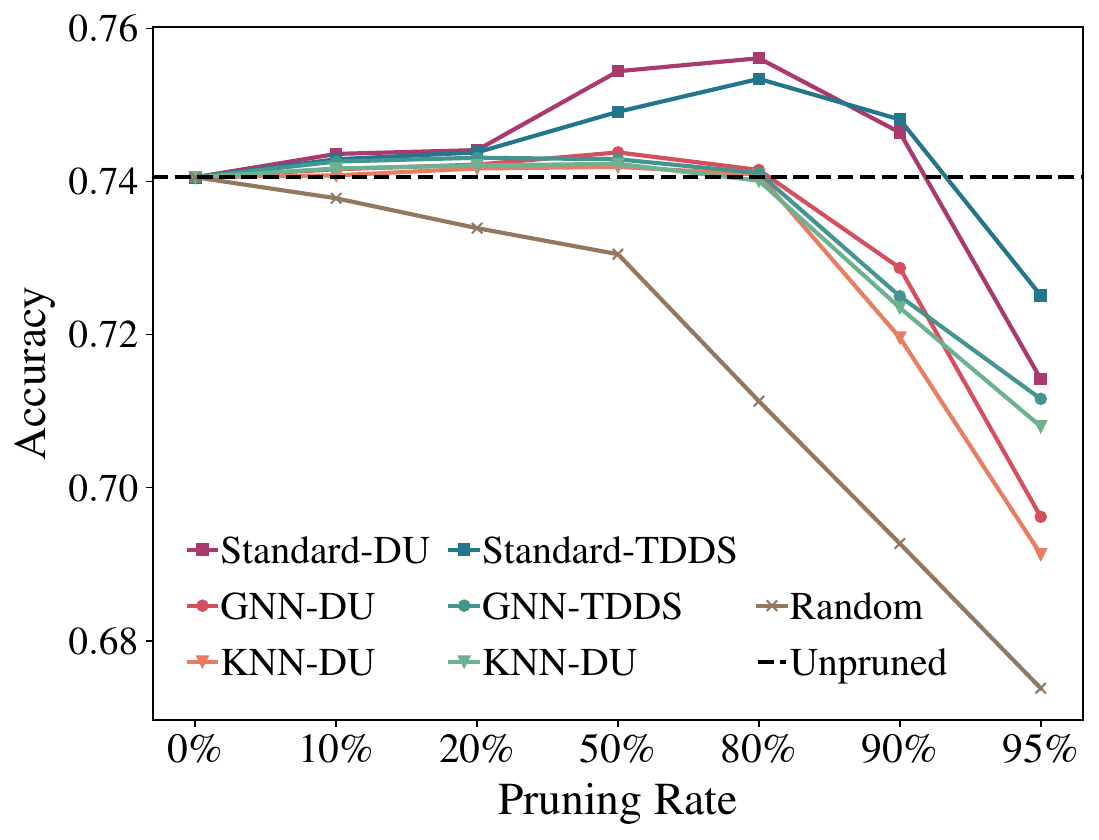}
}%
\subfloat[CIFAR-10\label{fig:pruningCifar10}]{%
 \includegraphics[width=0.49\textwidth]{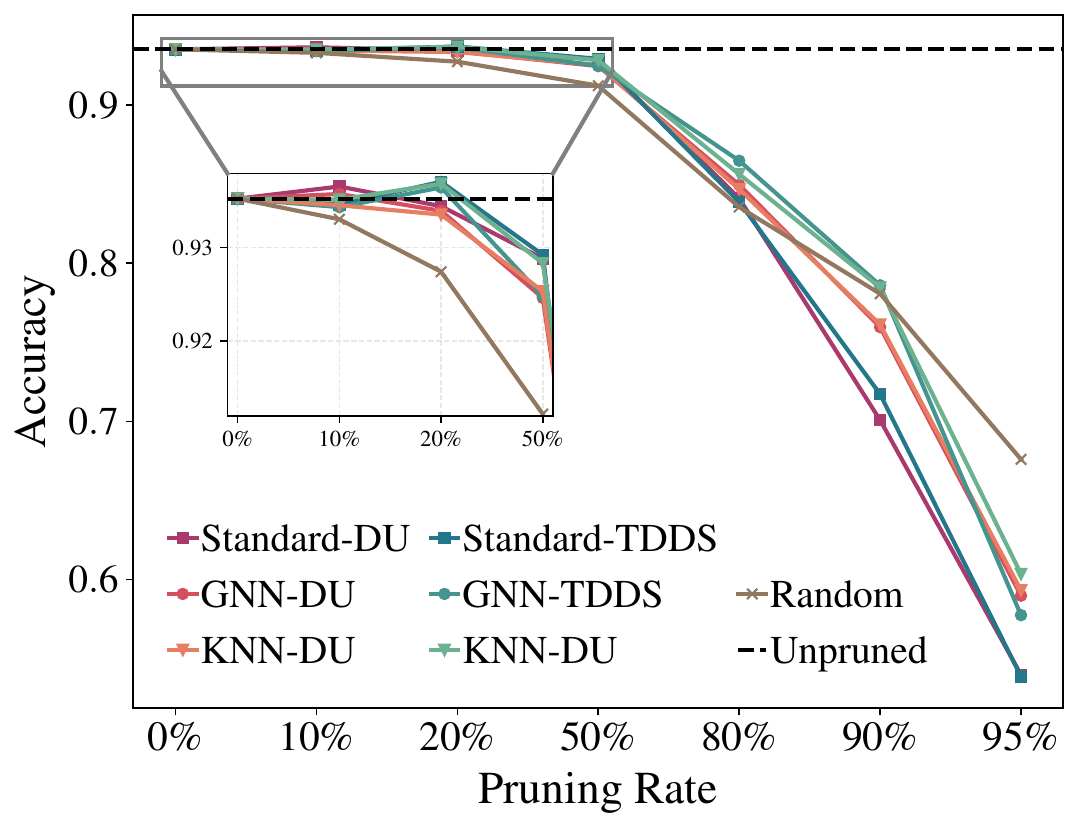}
}%
\caption{Evaluation of accuracy for different pruning rates and different datasets. Experiments are averaged over three seeds. In most setups, the GNN extrapolation scores outperform the KNN extrapolation. Especially for low pruning rates, the difference to the original pruning approach is quite low.}
\label{fig:secondExperiment}
\end{figure}

\textbf{Pruning Performance.} In \Cref{fig:secondExperiment} we compare pruning strategies at different pruning rates. The difference in final accuracy between ground truth scores to extrapolation methods is low for Places365 (\Cref{fig:pruningPlaces}) and ImageNet (\Cref{fig:pruningImageNet}). While TDDS maintains a higher accuracy for moderate pruning rates, the difference to our extrapolation decreases for higher pruning rates. DU performs worse than random pruning for pruning rates over $50\%$, making comparisons in this regime irrelevant. 

For synthetic CIFAR-100 1M, the original DU and TDDS actually improve model accuracy compared to standard training. We attribute this to the ability of the pruning methods to filter out noisy data. Similarly, score extrapolation techniques demonstrate performance gains over standard training, though the effect is less substantial than with direct pruning methods. 

For the smaller CIFAR-10 (\Cref{fig:pruningCifar10}), the performance between ground truth scores and extrapolated ones is identical for TDDS up to 20\% pruning. 
Importantly, across all datasets, extrapolated scores consistently outperform random pruning whenever the ground truth score does, demonstrating the practical value of score extrapolation.

\begin{figure}[bt]
\vskip -0.2cm
\centering
\subfloat[Places 365\label{fig:placesPareto}]{%
   \includegraphics[width=0.49\textwidth]{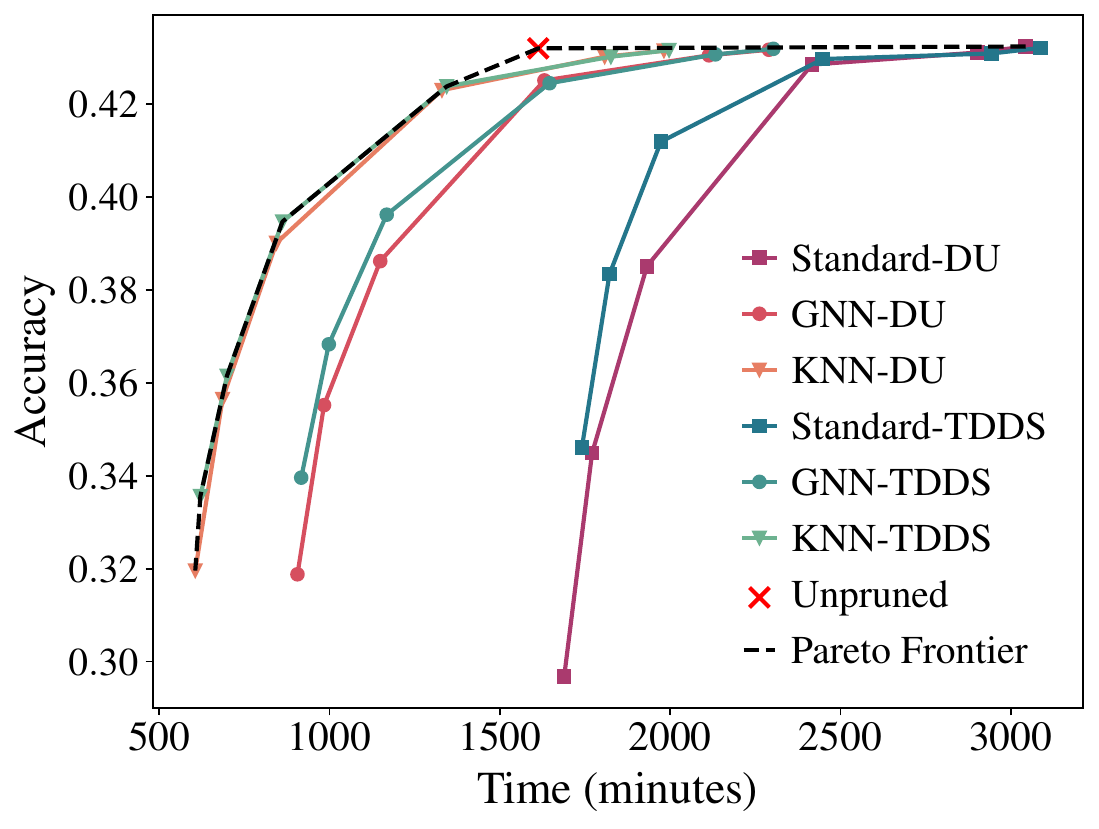}
}%
\subfloat[ImageNet\label{fig:ImageNetPareto}]{%
 \includegraphics[width=0.49\textwidth]{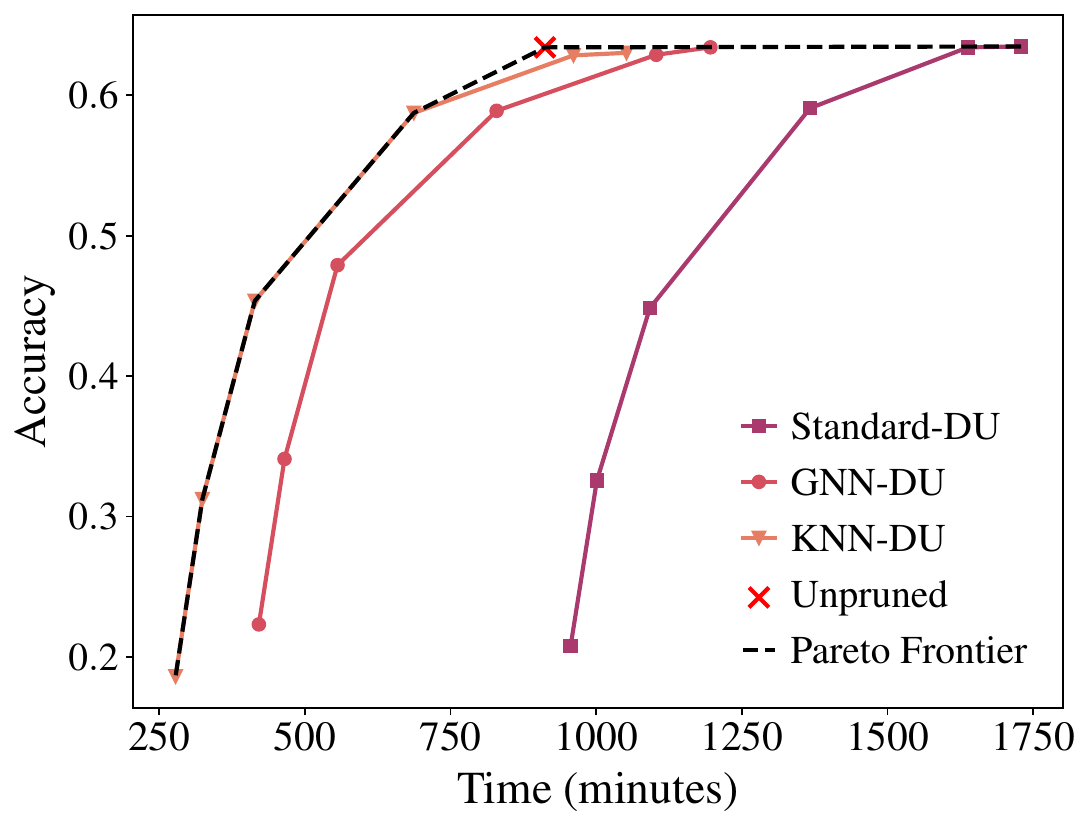}
}%
\vskip -0.2cm
\caption{Pareto plots of time to accuracy behavior for different pruning methods and the full dataset training. It can be seen that the pruning approaches are always slower than our extrapolation approach, while high pruning rates show faster than full training behavior.}
\label{fig:ParetoPlots}
\vskip -0.3cm
\end{figure}

\textbf{Time Optimality.}
We examine the computational behavior of the extrapolation approaches in more detail.
While the original scores demonstrate higher accuracy, indicating an upper bound for the extrapolation methods, their calculation is costly. \Cref{fig:ParetoPlots} compares the model performance with the pruning and training time.
We see that our extrapolated scores, especially the KNN extrapolation, are always Pareto optimal for Places365 and ImageNet. In contrast, the standard pruning approaches require more time for the standard pruning than even a regular full training. They would only obtain any real-time advantage if multiple unpruned training runs were to be performed, while our extrapolated scores provide a time advantage already at the first run.

\textbf{Relationship between Score Correlation and Downstream Task Performance.} 
To verify the results presented in the \Cref{tab:pruning-results}, we examine the correlation between the extrapolation and the original scores. \Cref{tab:correlation-results} reports Pearson \cite{benesty2009pearson} and Spearman \cite{zar2005spearman} correlation. As expected, the correlation increases with the subset size. We already saw that a 10\% subset size is sufficient for high performance in the pruning task in previously presented results. In addition, the GNN's correlation is always higher than the KNN's, underlining the GNN's greater ability to capture dataset properties.
\begin{table}[t]
  \centering
  \small
  \caption{Pearson and Spearman correlations for different pruning methods across datasets.
  GNN always achieves a higher correlation than KNN, and both approaches increase correlation with a higher training subset.}
  \vskip 0.2cm
  \label{tab:correlation-results}
  \renewcommand{\arraystretch}{1.1}
  \begin{tabular}{cccccccc}
    \toprule
    \multicolumn{1}{c}{\textbf{Dataset}} &
    \multicolumn{1}{c}{\textbf{Method}} &
    \multicolumn{1}{c}{\textbf{Subset size (\%)}} &
    \multicolumn{2}{c}{\textbf{Pearson $\rho$}} &
    \multicolumn{2}{c}{\textbf{Spearman $r_s$}} \\
    \cmidrule(lr){4-5} \cmidrule(lr){6-7}
    & & & \textbf{GNN} & \textbf{KNN} & \textbf{GNN} & \textbf{KNN} \\
    \midrule

    \multirow{2}{*}{Imagenet}
      & \multirow{2}{*}{DU}
      & 20 & \textbf{0.4193} & 0.3779 & \textbf{0.3503} & 0.3068 \\
      & 
      & 10 & \textbf{0.2850} & 0.2575 & \textbf{0.2178} & 0.1980 \\
    \cmidrule(lr){1-7}

    \multirow{4}{*}{Places365}
      & \multirow{2}{*}{DU}
      & 25 & \textbf{0.4004} & 0.3081 & \textbf{0.3611} & 0.2524 \\
      & 
      & 10 & \textbf{0.2612} & 0.2215 & \textbf{0.2158} & 0.1791 \\
      & \multirow{2}{*}{TDDS}
      & 25 & \textbf{0.2632} & 0.2251 & \textbf{0.2646} & 0.2214 \\
      & 
      & 10 & \textbf{0.2372} & 0.1620 & \textbf{0.2297} & 0.1594 \\
    \cmidrule(lr){1-7}

    \multirow{4}{*}{Synthetic CIFAR}
      & \multirow{2}{*}{DU}
      & 20 & \textbf{0.4910} & 0.4538 & \textbf{0.7009} & 0.6562 \\
      & 
      & 10 & \textbf{0.3396} & 0.3243 & \textbf{0.5593} & 0.5471 \\
      & \multirow{2}{*}{TDDS}
      & 20 & \textbf{0.4236} & 0.3955 & \textbf{0.6713} & 0.6244 \\
      & 
      & 10 & \textbf{0.3849} & 0.3273 & \textbf{0.5722} & 0.5324 \\
    \bottomrule
  \end{tabular}
  \vskip -0.3cm
\end{table}

 \begin{wraptable}{r}{7.0cm}
  \centering
  \small
  \vskip -0.45cm
  \caption{Pearson and Spearman correlations for unsupervised extrapolation for CIFAR10 with DU based on the Turtle~\cite{Gadetsky2024}. }
  \label{tab:unsupervised_results}
  \vskip -0.2cm
  \renewcommand{\arraystretch}{1.1}
  \begin{tabular}{ccccc}
    \toprule
    \multicolumn{1}{c}{\textbf{Subset size}} &
    \multicolumn{2}{c}{\textbf{Pearson $\rho$}} &
    \multicolumn{2}{c}{\textbf{Spearman $r_s$}} \\
    \cmidrule(lr){2-3} \cmidrule(lr){4-5}
    & \textbf{GNN} & \textbf{KNN} & \textbf{GNN} & \textbf{KNN} \\
    \midrule
    20\% & 0.4201 & \textbf{0.5481} & 0.5312 & \textbf{0.6630} \\
    10\% & 0.3692 & \textbf{0.5078} & 0.4519 & \textbf{0.6211} \\
    \bottomrule
  \end{tabular}
  \vskip -0.4cm
\end{wraptable}

\begin{figure}[tb]
    \centering
    \includegraphics[width=\textwidth]{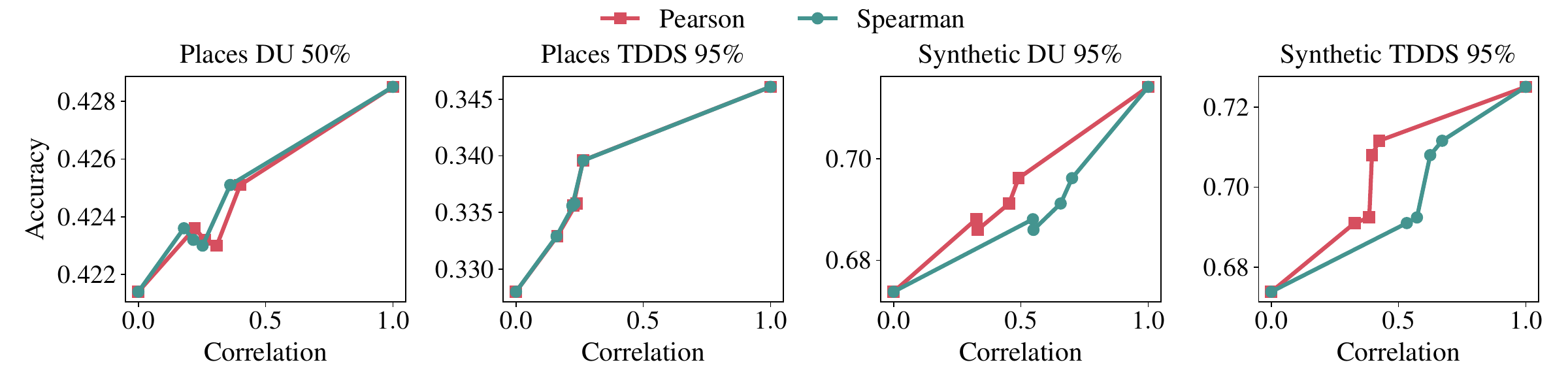}
    \caption{Analysis of the dependency of correlation and accuracy of the extrapolation methods for Places365 and synthetic CIFAR-100. The accuracy increases with the correlation of the specific task.}
    \label{fig:CorrToAcc}
    \vskip -0.5cm
\end{figure}

As the next step, we investigated the actual relationship of these correlation scores to the downstream task performance. In \Cref{fig:CorrToAcc}, we show the Pearson and Spearman correlation over the accuracy. The correlation of the original and extrapolated scores perfectly aligned with achieved accuracy, confirming the obtained pruning results of our extrapolated scores. 

\textbf{Unsupervised Data Pruning.}
Since our extrapolation paradigm works effectively for classic data pruning, we modify the labeled data constraints of data pruning to investigate the flexibility and scalability of our approach. When dealing with large datasets, the assumption that all data is labeled might not be realistic. Thus, we modify the original data pruning task to work with unlabeled datasets. We use Turtle~\cite{Gadetsky2024} for the unsupervised image classification on CIFAR-10, and applied the score calculation with DU and our extrapolations. Further experiment details can be found in \Cref{ap:experimentSetup}.

For our newly created unsupervised data pruning setup, we focus on analyzing the correlation of our extrapolation approaches to the original DU score. 
In \Cref{tab:unsupervised_results} we can see that the correlations of the extrapolated scores are as high as for the standard pruning approach, indicating the flexibility of the extrapolation paradigm. Interestingly, KNN has a higher correlation than GNN. Which might be caused by the foundation model defined latent space that Turtle uses. In addition, the correlation increases with the size of the subset.

\textbf{Adversarial training.}
In addition to the supervised and unsupervised, we investigate the adversarial training scenario, where our extrapolation can confirm its high effectiveness in selecting a subset for adversarial training. We perform adversarial training in the $\ell_{\infty}$-norm with a perturbation budget of $\epsilon=8/255$ on the CIFAR-10 \cite{cifar} dataset. We use the same training hyperparameters as in~\cite{wang2023better}. In each case, we prune 25\% of the data. 
Table~\ref{tbl:extrapolated_vs_GT_adv} summarizes the robustness for the different methods. Score extrapolation considerably improves upon random pruning in terms of robustness and model accuracy on clean data. As expected, it performs only slightly worse than using ground truth scores directly. The KNN extrapolation approach achieves $0.54$ linear correlation with the ground truth scores, which could be improved by more sophisticated extrapolation approaches.

In Table~\ref{tbl:extrapolation_synthetic_adv_results}, we additionally provide results for KNN-based extrapolation for a larger dataset. Here, we also evaluate our approach for $\ell_2$-based adversarial training ($\epsilon = 128/255$). We extrapolate scores from the DU scores from a standard CIFAR-10 training run to $2$ million synthetic CIFAR-10 images from~\cite{wang2023better} and prune $50\%$ of the data. We do not compare to pruning with ground truth scores, as performing a full adversarial training run with $2$ million data samples was too expensive. Score extrapolation outperforms random pruning in both settings, while introducing only negligible computational overhead ($<6\%$). 
This result demonstrates the effectiveness of score extrapolation in adversarial training, when the initial seed dataset is considerably smaller than the full dataset (i.e., $m << n$).

\begin{minipage}[t]{0.48\textwidth}
\centering
\captionof{table}{Comparison of the utility of original and extrapolated DU scores on the CIFAR-10 dataset for \(\ell_{\infty}\)-norm adversarial training (\(\epsilon=8/255\)). $25\%$ of the samples are pruned for each method. Random pruning is provided as a baseline.}
\label{tbl:extrapolated_vs_GT_adv}
\vspace{0.2cm}
\begin{small}
\begin{tabular}{lll}
\toprule
Experiment & Robust & Clean \\
\midrule
Random & 47.95\% & 80.43\% \\
Extrapolated-KNN & 50.39\% & 80.86\% \\
Ground Truth Scores & 52.26\% & 81.95\% \\
\bottomrule
\end{tabular}
\end{small}
\vspace{0.2cm}
\end{minipage}
\hfill
\begin{minipage}[t]{0.48\textwidth}
\centering
\captionof{table}{Evaluation of extrapolated DU scores on a $2$ million sample synthetic CIFAR-10 dataset. $50\%$ of the $2$ million synthetic samples are pruned for each method. We provide random pruning as a baseline.}
\label{tbl:extrapolation_synthetic_adv_results}
\vspace{0.2cm}
\begin{small}
\begin{sc}
\resizebox{\textwidth}{!}{
\begin{tabular}{llll}
\toprule
Norm & Experiment & Robust & Clean \\
\midrule
\multirow{2}{*}{\(\ell_2\)} 
& Random & 80.79\% & 94.20\% \\
& Extrapolated KNN & 81.28\% & 94.22\% \\
\midrule
\multirow{2}{*}{\(\ell_\infty\)} 
& Random & 63.13\% & 90.86\% \\
& Extrapolated & \textbf{63.56}\% & 90.50\% \\
\bottomrule
\end{tabular}
}
\end{sc}
\end{small}
\vspace{0.2cm}
\end{minipage}

\begin{figure}
\centering
\vskip -0.2cm
\begin{tabular}{cc}
\adjustbox{valign=b}{\subfloat[Correlation of Scores\label{subfig-1:dummy}]{%
      \includegraphics[width=.46\linewidth]{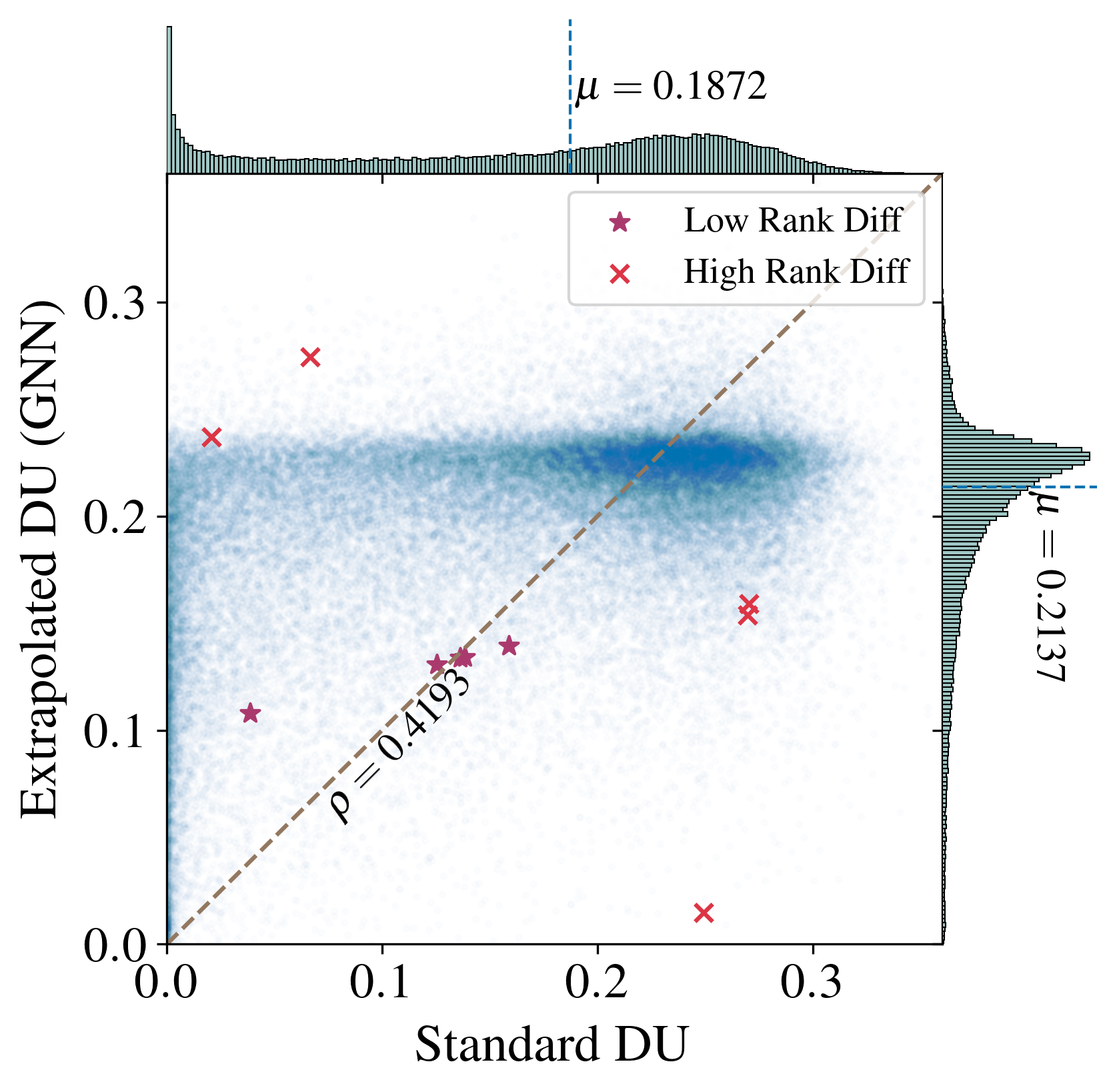}}}
&      
\adjustbox{valign=b}{\begin{tabular}{@{}c@{}}
\subfloat[High rank difference examples\label{subfig-2:dummy}]{%
      \includegraphics[width=.52\linewidth]{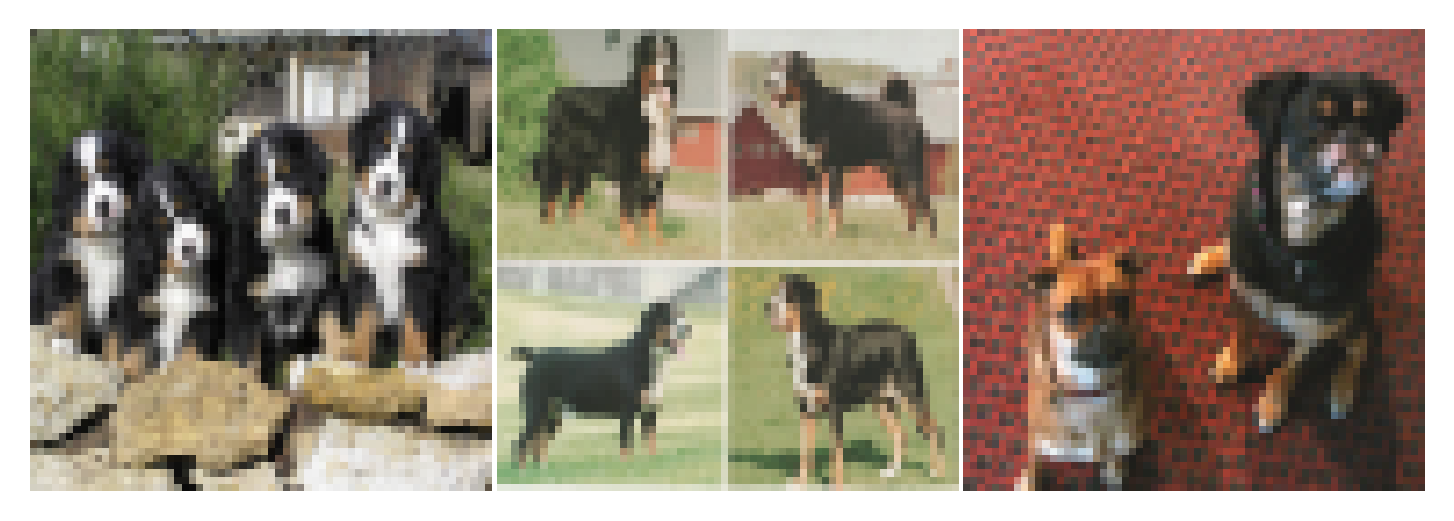}} \\
\subfloat[Low rank difference examples\label{subfig-3:dummy}]{%
      \includegraphics[width=.52\linewidth]{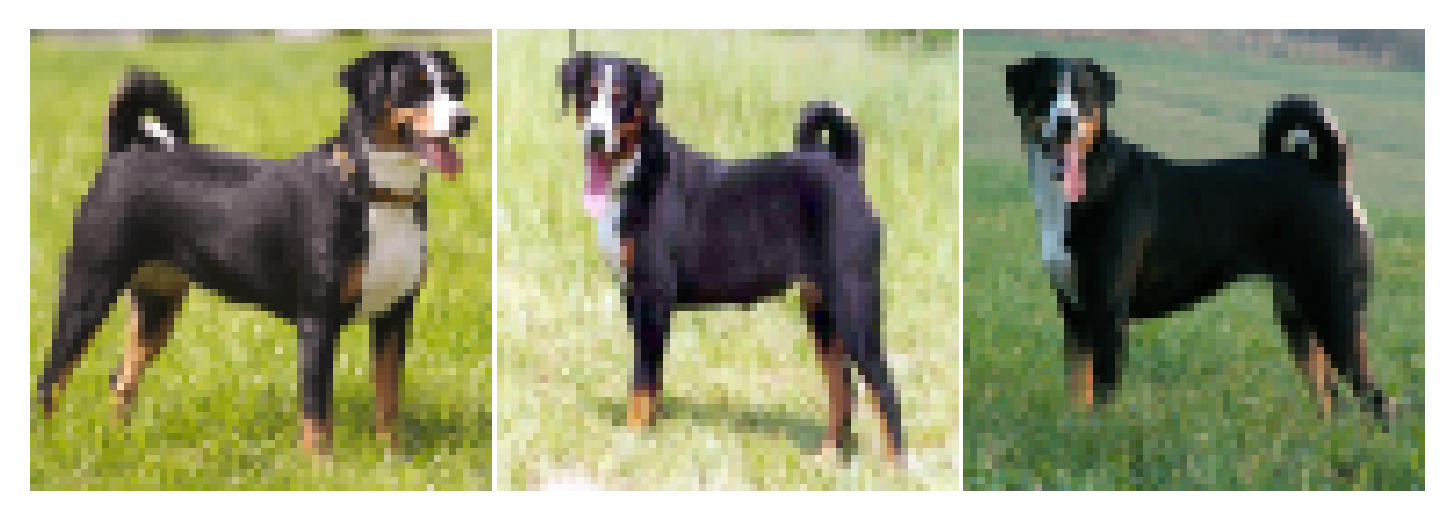}}
\end{tabular}}
\end{tabular}
\caption{Score distribution (a) and qualitative analysis (b-c) of extrapolation errors for ImageNet and DU. a) Extrapolated scores show a moderate correlation with ground truth but miss the bimodal structure, resulting in a narrower, oversmoothed distribution. (b) and (c) show examples with high and low rank discrepancies. High discrepancies often correspond to outliers with atypical backgrounds or multiple objects, while low discrepancies align with prototypical class examples.}\label{fig:VisualAndDistribution}
\vskip -0.2cm
\end{figure}

\textbf{Limitations and Visual Analysis.} In \Cref{fig:VisualAndDistribution} (a), we perform a deeper investigation regarding limitations of our extrapolation approach, superficially focusing on extrapolated DU scores on the ImageNet dataset. While the extrapolation method achieves a moderate correlation between ground truth and estimated scores, the extrapolated scores fail to capture the bimodal structure present in the original distribution, instead forming a narrower, unimodal distribution with a higher mean and lower variance. This oversmoothing likely contributes to the observed extrapolation errors. 

To further investigate these errors, we highlight samples with high and low rank differences (specifically using one "dog" class, but other classes showed similar patterns). Qualitative examples in subfigures (b) and (c) suggest that high rank differences often correspond to atypical or out-of-distribution samples—such as those with unusual backgrounds, multiple subjects, or low visual quality—whereas low rank difference samples tend to be prototypical, centered, and consistent in appearance. This indicates that the extrapolation method struggles most with outliers and visually ambiguous inputs. We hypothesize that more powerful extrapolation methods or using more seed data to train the score extrapolation would improve score extrapolation in these cases.

\section{Conclusion}
Since data pruning usually requires a full training on the dataset to estimate the scores for a subset selection, its applicability to large-scale datasets and real-world applications is limited. To mitigate this problem, we propose a novel score extrapolation paradigm. Instead of training on the full set, we select a small subset to estimate the initial scores for the chosen data pruning method and extrapolate these scores to the remaining samples in the dataset using a computationally efficient extrapolation method. In this work, we propose a KNN- and a GNN-based extrapolation scheme, which are easily applicable to different learning tasks. Our experiments show that the extrapolated scores show high correlation with the original scores and achieve a high downstream task performance for two different pruning scores, three different tasks, and four different datasets. A closer look at the extrapolation errors shows that our initial methods fail to capture the full complexity of the ground truth score distributions. A visual investigation reveals that extrapolation is particularly challenging for outlier samples that deviate from prototypical class examples.

\textbf{Outlook.} In future work, we aim to improve downstream task performance, which could include increasing extrapolation accuracy, for example, by refining the GNN-based score estimator, exploring new extrapolation methods, and investigating alternative subset selection strategies. 
Moreover, we aim to extend our extrapolation approach to other data selection tasks involving costly score computations, such as influence functions~\cite{koh2017understanding} and data attribution~\cite{ilyas2022datamodels}. 
Finally, we believe score extrapolation could also benefit tasks beyond data selection, such as out-of-distribution detection.

\begin{ack}
This research was supported and partially funded by Pontificia Universidad Católica de Valparaíso, Vice Rector's Office for Research, Creation and Innovation within the programs "DI Iniciación" (039.485/2024) and "Semilla 2023" (039.247)

\end{ack}

\clearpage
\appendix

\section{Experiment Setup}
\label{ap:experimentSetup}

\subsection{Dataset}
\label{ap:dataset and Assets}
To evaluate the efficacy of our proposed score extrapolation framework, we conduct experiments on four image classification datasets differing in scale, number of classes, and image resolutions: CIFAR-10 \cite{cifar} (50K samples, $10$ classes, $32 \times 32$), synthetic CIFAR-100 \cite{wang2023better} (1M samples, $100$ classes, $32 \times 32$, generated with denioising diffusion models \cite{ho2020denoising}), Places-365 \cite{zhou2017places} (1.80M samples, $365$ classes, resized to $64 \times 64$ to expedite experiments), and Imagenet-1k \cite{Deng2009} (1.28M samples, $1000$ classes, downscaled version of $64 \times 64$). To evaluate model accuracy after pruning, we use the original test sets of each dataset, with the exception of synthetic CIFAR-100, for which we employ the standard CIFAR-100 \cite{cifar} test set. Additionally, we explore the unsupervised dataset pruning with standard CIFAR-10, and adversarial training with synthetic CIFAR-10 ~\cite{wang2023better} (2M samples).

\textbf{Dataset License:}
All datasets used in our experiments are publicly available, and most of them are widely used in the ML community. The standard CIFAR-10 and CIFAR-100~\cite{cifar} datasets are freely available for research and educational purposes without any licensing requirements. ImageNet ~\cite{Deng2009} is available for free to researchers for non-commercial use but does not outline a specific license. Both synthetic CIFAR-10 and synthetic CIFAR-100~\cite{wang2023better}\footnote{\url{https://github.com/wzekai99/DM-Improves-AT/}} are publicly available under the MIT license. Similarly, Places365~\cite{zhou2017places} is released under the MIT license. We performed experiments adhering to the licensing terms of the respective datasets.

\subsection{Statistical Significance}
\label{ap:statistical_significance}
Since the computed scores depend on the training, which is stochastic by nature, the scores obtained at the end are also stochastic. To ensure statistical robustness, for any dataset, we compute three different sets of ground truth scores $S \in \mathbb{R}^n$ with three different random initializations. For each set of scores, we compute the model accuracy at various pruning rates: $10, 20, 50, 80, 90$ and  $95$ percentages. To compute subset scores $S_s$, we follow the same procedure. We randomly select $\sD_s \subset \sD$ of cardinality $m$, and compute the scores $S_s$ with standard pruning algorithms. We do this three times with three different seeds. For each set of $S_s$, we extrapolate using the KNN and GNN approaches. In this way, we obtain six sets of extrapolated scores $S_r$, (three from KNN, and three from GNN). Note that extrapolation with GNN itself is stochastic in nature. For each set of scores, we prune the data at various rates and train the model on the pruned dataset once. Thus, for each pruning rate, we get three test accuracy values for both extrapolation methods. \Cref{fig:secondExperiment}, \Cref{fig:appendix_pruning} and \Cref{fig:ParetoPlots} report this mean accuracy.

\subsection{Computational Resources}
\label{ap:comutational_resources}
All experiments are conducted using NVIDIA A100-PCIE GPUs, with 42.4 GB of VRAM. Computational time reported in \Cref{tab:pruning-results}, and \Cref{fig:ParetoPlots} are the total mean runtime (in minutes) required to compute $S_s$, extrapolate, and subsequent model training on the pruned dataset. Time for ground truth scores reflects the mean time required for full dataset scoring plus training on the pruned subset.

\subsection{Models}
\label{ap:Models}
Both DU, and TDDS require model training for numerous epochs to compute the scores. To validate that score extrapolation works with different models $\mathcal{F}_s$, we used ResNet-18 \cite{he2016deep} for Cifar-10, and Imagenet, and ResNet-50 for Synthetic CIFAR-100, and Places-365. For the adversarial setting, we used Wide-ResNet-28-10 \cite{zagoruyko2016wide}. During extrapolation, samples are represented in the embedding space induced by these models.

For the unsupervised setting, we employ DINOv2~\cite{oquab2023dinov2} as a foundation model to obtain fixed embeddings for all samples. Both extrapolation procedures (KNN-based and GNN-based) are subsequently performed in the embedding space of this foundation model. This diversity of architectures and training paradigms demonstrates that our extrapolation approach is not restricted to a specific model but can be applied broadly across a range of backbone networks and training schemes.

\subsection{Experiment Hyperparameters}
\label{sec:exp_hyperparams}
We collected all hyperparameter setting in\Cref{tab:hyperparams_unsupervised,tab:hyperparams_tdds,tab:hyperparams_pruning,tab:hyperparams_du}. They are properly introduced with the scored description in the following section.

\section{Scores}
\label{ap:Scores}
We assess our extrapolation framework with two state-of-the-art dataset pruning methods: Dynamic Uncertainty (DU) \cite{he2024large}, and Temporal Dual-Depth Scoring (TDDS) \cite{zhang2024spanning}.

\subsection{DU Scores} Given a model $\theta_k$ trained over $K$ epochs, $U_k(x)$ for a sample $x$ at epoch $k$ is computed as the standard deviation of the predicted probabilities $\mathbb{P}(y|x, \theta_k)$ over a sliding window of $J$ epochs \cite{he2024large}:

\[
S_k(x) = \sqrt{\frac{1}{J-1} \sum_{j=0}^{J-1} \left[ \mathbb{P}(y|x, \theta_{k+j}) - \bar{\mathbb{P}} \right]^2},
\]
where $\bar{\mathbb{P}} = \frac{1}{J} \sum_{j=0}^{J-1} \mathbb{P}(y|x, \theta_{k+j})$. The final dynamic uncertainty score $S(x)$ for each sample is computed by averaging over all sliding windows:

\[
S(x) = \frac{1}{K-J} \sum_{k=0}^{K-J-1} S_k(x),
\]

For experiments, we set $J = 10$, and $K = 50$ for CIFAR-10, synthetic CIFAR-100 and PLACES-365, while $K = 90$ is used for Imagenet. More details on the hyperparameters are provided in \Cref{tab:hyperparams_du}.

\subsection{TDDS Scores} 
TDDS \cite{zhang2024spanning} computes the importance score for a sample $x$ by quantifying its contribution to optimization dynamics. Specifically, TDDS calculates the epoch-wise change in loss, $\Delta \ell_k(x)$, projected onto the model's optimization trajectory. Formally, for a sliding window of size $K$, the score is computed as:

$$
S(x) = \sum_{k=J}^{K}\beta (1-\beta)^{K-k}\sum_{j=k-J+1}^{k}\left(\left|\Delta \ell_j(x)\right| - \frac{1}{J}\sum_{i=k-J+1}^{k}\left|\Delta \ell_i(x)\right|\right)^2,
$$

where $\Delta\ell_k(x)$ measures the KL-divergence of predictions between consecutive epochs:

$$
\Delta \ell_k(x) = f_{\theta_{k+1}}(x)^\top \log\frac{f_{\theta_{k+1}}(x)}{f_{\theta_k}(x)},
$$

and $\beta$ is an exponential decay factor. In experiments, we set $J=10$, and $\beta=0.9$ for all datasets, and $K=50$ for CIFAR-10, synthetic CIFAR-100, and Places-365, whereas $K=90$ for ImageNet. Further details are provided in \Cref{tab:hyperparams_tdds}.

\subsection{Unsupervised DU Scores}
\label{sec:unsup_du}
To assess whether our score-extrapolation framework remains effective in the absence of ground-truth labels, we employ TURTLE~\cite{Gadetsky2024}. TURTLE assigns a pseudo-label to each sample $x \in \mathcal{D}$ by optimizing a bilevel objective within the representation space induced by a foundation model $\phi$.

During TURTLE optimization, at each outer iteration $k \in \{1, \ldots, K\}$, we record the softmax probability vector:
\[
    \mathbf{p}^{(k)}(x) = \operatorname{softmax}\left(\boldsymbol{\theta}^{(k)} \phi(x)\right) \in \Delta^{C-1},
\]
where $\boldsymbol{\theta}^{(k)}$ denotes the learnable linear transformation at iteration $k$, and $C$ is the number of classes specified a priori. After $K$ outer iterations, we define the final pseudo-label for $x$ as
\[
    \hat{y}(x) = \arg\max_{c \in \{1, \ldots, C\}} \mathbf{p}^{(K)}_c(x).
\]
Analogous to the supervised setting, we perform post-hoc computation to obtain the uncertainty at the pseudo-label $\hat{y}(x)$ across a sliding window of length $J$ over epochs and subsequently average these values to compute the \textit{unsupervised-DU} score for each sample. 

Similar to the supervised setting, we use $J = 10$ for the experiments. Further hyperparameter details are provided in \Cref{tab:hyperparams_unsupervised}, which follows settings in \cite{zhang2024spanning}.

\begin{table}[h]
\centering
\caption{Hyperparameters and experimental settings for all datasets to compute standard DU scores $\sS$, as well as subset DU scores $\sS_s$. Subset sizes are reported as a percentage of the total dataset size. To compute standard scores, training is done on the complete dataset}
\label{tab:hyperparams_du}
\begin{tabular}{lcccc}
\toprule
\textbf{Hyperparameters} & \textbf{CIFAR-10} & \textbf{Synthetic CIFAR-100} & \textbf{Places-365} & \textbf{ImageNet} \\
\midrule
Num epochs ($K$) & $50$ & $50$ & $50$ & $90$ \\
Batch size ($B$) & $256$ & $256$ & $128$ & $256$ \\
Model ($\mathcal{F}$) & ResNet-18 \cite{he2016deep} & ResNet-50 & ResNet-50 & ResNet-18 \\
Optimizer & Adam \cite{kingma2014adam} & Adam & Adam & Adam \\
Learning rate ($\eta$) & $10^{-3}$ & $10^{-3}$ & $10^{-3}$ & $10^{-3}$ \\
Weight decay ($\lambda$) & $10^{-4}$ & $10^{-4}$ & $10^{-4}$ & $10^{-4}$ \\
Scheduler & None & None & None & None \\
Window ($J$) & $10$ & $10$ & $10$ & $10$ \\
Subset size ($m$) & $40\%$, $20\%$ & $30\%$, $20\%$, $10\%$ & $25\%$, $10\%$ & $20\%$, $10\%$ \\
\bottomrule
\end{tabular}
\end{table}

\begin{table}[h]
\centering
\caption{Hyperparameters to compute standard and subset TDDS scores.}
\label{tab:hyperparams_tdds}
\begin{tabular}{lcccc}
\toprule
\textbf{Hyperparameters} & \textbf{CIFAR-10} & \textbf{Synthetic CIFAR-100} & \textbf{Places-365} \\
\midrule
Num epochs ($K$) & $50$ & $50$ & $50$ \\
Batch size ($B$) & $256$ & $256$ & $128$ \\
Model ($\mathcal{F}$) & ResNet-18 & ResNet-50 & ResNet-50  \\
Optimizer & SGD & SGD & SGD  \\
Learning rate ($\eta$) & $10^{-3}$ & $10^{-3}$ & $10^{-3}$ \\
Weight decay ($\lambda$) & $5 \times 10^{-4}$ & $5 \times 10^{-4}$ & $5 \times 10^{-4}$ \\
Momentum & $0.9$ & $0.9$ & $0.9$  \\
Nesterov \cite{sutskever2013importance} & True & True & True  \\
Scheduler & CosineAnnealing \cite{loshchilov2016sgdr} & CosineAnnealing & CosineAnnealing \\
Window ($J$) & $10$ & $10$ & $10$ \\
Trajectory & $10$ & $10$ & $10$ \\
Exponential decay ($\beta$) & $0.9$ & $0.9$ & $0.9$ \\
Subset size ($m$) & $40\%$, $20\%$ & $20\%$, $10\%$ & $25\%$, $10\%$ \\
\bottomrule
\end{tabular}
\end{table}

\begin{table}[h]
\centering
\caption{Hyperparameters used for training models on pruned datasets. Both random pruning, and score based pruning (standard as well as extrapolated scores) use the same configurations}
\label{tab:hyperparams_pruning}
\begin{tabular}{lcccc}
\toprule
\textbf{Hyperparameters} & \textbf{CIFAR-10} & \textbf{Synthetic CIFAR-100} & \textbf{Places-365} & \textbf{ImageNet} \\
\midrule
Num epochs ($K$) & $50$ & $50$ & $50$ & $90$ \\
Batch size ($B$) & $256$ & $256$ & $128$ & $256$ \\
Model ($\mathcal{F}$) & ResNet-18 & ResNet-50 & ResNet-50 & ResNet-18 \\
Optimizer & Adam & Adam & Adam & Adam \\
Learning rate ($\eta$) & $10^{-3}$ & $10^{-3}$ & $10^{-3}$ & $10^{-3}$ \\
Weight decay ($\lambda$) & $10^{-4}$ & $10^{-4}$ & $10^{-4}$ & $10^{-4}$ \\
Scheduler & OneCycle \cite{smith2019super} & OneCycle & OneCycle & OneCycle \\
Window ($J$) & $10$ & $10$ & $10$ & $10$ \\
\bottomrule
\end{tabular}
\end{table}

\begin{table}[h]
\centering
\caption{Hyperparameters to compute unsupervised DU scores (for full dataset, as well as subset)}
\label{tab:hyperparams_unsupervised}
\begin{tabular}{lcccc}
\toprule
\textbf{Hyperparameters} & \textbf{CIFAR-10}\\
\midrule
Num epochs ($K$) & $400$  \\
Batch size ($B$) & $10,000$  \\
Inner steps ($M$) & $10$ \\
Representation ($\phi$) &  DINOv2~\cite{oquab2023dinov2}  \\
Optimizer & Adam  \\
Regularization coeff. ($\gamma$) & $10$  \\
Inner Learning rate ($\eta_{\text{in}}$) & $10^{-3}$ \\
Outer Learning rate ($\eta_{\text{out}}$) & $10^{-3}$ \\
Weight decay ($\lambda$) & $10^{-3}$  \\
Scheduler & None  \\
Window ($J$) & $10$ \\
Subset size ($m$) & 40\%  \\
\bottomrule
\end{tabular}
\end{table}

\subsection{Scores Extrapolation}
For KNN-based extrapolation, we computed the $k$ nearest neighbors using the Euclidean distance. To assess how the choice of $k$ affects extrapolation, we varied $k$ across ${10, 20, 50, 100}$ and evaluated the correlation between the extrapolated and ground-truth scores (based on $S)$ for samples in $\sD_r$. The value of $k$, yielding the highest Pearson correlation, is reported in our main results (\Cref{tab:correlation-results}), while the full ablation is presented in \Cref{tab:all_correlation_knn}.

For GNN-based extrapolation, we similarly examined the effect of the neighborhood size $k$ while constructing the graph. We used different values of $k$ $(10, 20,$ and $50)$. The GNN comprises three GCN~\cite{kipf2016semi} layers (hidden dimensions $512$ and $256$) and an output layer producing scalar importance scores. We use dropout regularization of $0.5$ to avoid overfitting. To ensure scalability on large dataset, we use neighbor sampling \cite{hamilton2017inductive}, with mini-batches of nodes size $128$. 

We randomly split $m$ samples in $\sD_s$ into $90\%$ training and $10\%$ validation set. These nodes already have the computed scores $S_s$. We train GNN for $25$ epochs, and the checkpoint achieving the highest Pearson correlation between predicted and reference scores (based on $S_s$) on the validation set is selected for inference. Scores for all samples in $\sD_r = \sD \setminus \sD_s$ are inferred using this model checkpoint. We report the correlation between the inferred scores and scores with the standard approach ($S$) for the samples in $\sD_r$.

Interestingly, we observe that GNNs usually achieved the best performance with smaller neighborhood sizes ($k=10$), suggesting that message passing enables effective propagation of information even with sparse local connectivity. Detailed results are provided in \Cref{tab:all_correlation_gnn}, with the best-performing configuration reported in \Cref{tab:correlation-results}.  

\begin{table}[t]
  \centering
  \small
  \setlength{\tabcolsep}{3pt} %
  \caption{Pearson and Spearman correlations for different $k$ in KNN pruning methods across datasets.}
  \label{tab:all_correlation_knn}
  \renewcommand{\arraystretch}{0.9}
  \begin{tabular}{cccccccccccc}
    \toprule
    \multicolumn{1}{c}{\textbf{Dataset}} &
    \multicolumn{1}{c}{\textbf{Method}} &
    \multicolumn{1}{c}{\textbf{Sample size (\%)}} &
    \multicolumn{4}{c}{\textbf{Pearson $\rho$}} &
    \multicolumn{4}{c}{\textbf{Spearman $r_s$}} \\
    \cmidrule(lr){4-7} \cmidrule(lr){8-11}
    & & & $k{=}10$ & $k{=}20$ & $k{=}50$ & $k{=}100$ & $k{=}10$ & $k{=}20$ & $k{=}50$ & $k{=}100$ \\
    \midrule

    \multirow{2}{*}{Imagenet}
      & \multirow{2}{*}{DU}
      & 20 & $0.3764$ & $\textbf{0.3779}$ & $0.3698$ & $0.3607$ & $0.2911$ & $\textbf{0.3068}$ & $0.3057$ & $0.2982$ \\
      & 
      & 10 & $0.2507$ & $0.2572$ & $\textbf{0.2575}$ & $0.2513$ & $0.1773$ & $0.1888$ & $\textbf{0.1980}$ & $0.1894$ \\
    \cmidrule(lr){1-11}

    \multirow{4}{*}{Places365}
      & \multirow{2}{*}{DU}
      & 25 & $0.2904$ & $0.3054$ & $\textbf{0.3081}$ & $0.3004$ & $0.2283$ & $0.2446$ & $\textbf{0.2524}$ & $0.2487$ \\
      & 
      & 10 & $0.2139$ & $0.2196$ & $\textbf{0.2215}$ & $0.2184$ & $0.1687$ & $0.1722$ & $\textbf{0.1791}$ & $0.1715$ \\
      & \multirow{2}{*}{TDDS}
      & 25 & $0.2169$ & $0.2208$ & $\textbf{0.2251}$ & $0.2213$ & $0.2128$ & $0.2180$ & $\textbf{0.2214}$ & $0.2159$ \\
      & 
      & 10 & $0.1497$ & $0.1561$ & $\textbf{0.1620}$ & $0.1583$ & $0.1438$ & $0.1526$ & $\textbf{0.1594}$ & $0.1513$ \\
    \cmidrule(lr){1-11}

    \multirow{4}{*}{Synthetic}
      & \multirow{3}{*}{DU}
      & 30 & $0.4964$ & $0.5092$ & $\textbf{0.5118}$ & $0.5106$ & $0.6675$ & $0.6864$ & $\textbf{0.6962}$ & $0.6955$ \\
      & 
      & 20 & $0.4447$ & $0.4530$ & $\textbf{0.4538}$ & $0.4499$ & $0.6355$ & $0.6493$ & $\textbf{0.6562}$ & $0.6562$ \\
      & 
      & 10 & $0.3123$ & $0.3197$ & $\textbf{0.3243}$ & $0.3202$ & $0.5286$ & $0.5374$ & $\textbf{0.5471}$ & $0.5438$ \\
      & \multirow{2}{*}{TDDS}
      & 20 & $0.3821$ & $0.3886$ & $\textbf{0.3955}$ & $0.3910$ & $0.6067$ & $0.6122$ & $\textbf{0.6208}$ & $0.6197$ \\
      & 
      & 10 & $0.3096$ & $0.3178$ & $\textbf{0.3273}$ & $0.3147$ & $0.5137$ & $0.5261$ & $\textbf{0.5324}$ & $0.5311$ \\
    \cmidrule(lr){1-11}

    \multirow{4}{*}{CIFAR-10}
      & \multirow{2}{*}{DU} 
      & 40 & $0.6149$ & $0.6328$ & $\textbf{0.6371}$ & $0.6313$ & $0.6225$ & $0.6411$ & $\textbf{0.6477}$ & $0.6447$ \\
      &
      & 20 & $0.2721$ & $0.2759$ & $\textbf{0.2765}$ & $0.2762$ & $\textbf{0.2603}$ & $0.2597$ & $0.2558$ & $0.2507$ \\
      & \multirow{2}{*}{TDDS} 
      & 40 & $0.3841$ & $0.4126$ & $\textbf{0.4181}$ & $0.4158$ & $0.5593$ & $0.5898$ & $0.6029$  & $\textbf{0.6040}$ \\
      &
      & 20 & $0.2014$ & $0.2128$ & $\textbf{0.2134}$ & $0.2038$ & $0.2738$ & $0.2806$ & $\textbf{0.2764}$  & $0.2611$ \\
    \bottomrule
  \end{tabular}
\end{table}

\begin{table}[t]
  \centering
  \setlength{\tabcolsep}{4pt} %
  \caption{Pearson and Spearman correlations for different $k$ in GNN based extrapolation}
  \label{tab:all_correlation_gnn}
  \renewcommand{\arraystretch}{0.9}
  \begin{tabular}{cccccccccc}
 \toprule
 \textbf{Dataset} & \textbf{Method} & \textbf{Sample size (\%)} &
 \multicolumn{3}{c}{\textbf{Pearson $\rho$}} &
 \multicolumn{3}{c}{\textbf{Spearman $r_s$}} \\
 \cmidrule(lr){4-6} \cmidrule(lr){7-9}
 & & & $k{=}10$ & $k{=}20$ & $k{=}50$ & $k{=}10$ & $k{=}20$ & $k{=}50$ \\

    \midrule

    \multirow{2}{*}{Imagenet}
      & \multirow{2}{*}{DU}
      & 20 & $\textbf{0.4193}$ & $0.4139$ & $0.4045$ & $\textbf{0.3503}$ & $0.3442$ & $0.3388$  \\
      & 
      & 10 & $\textbf{0.2850}$ & $0.2734$ & $0.2784$ & $0.2178$ &  $\textbf{0.2189}$ & $0.2159$ \\
    \cmidrule(lr){1-9}

    \multirow{4}{*}{Places365}
      & \multirow{2}{*}{DU}
      & 25  & $\textbf{0.4004}$ & $0.3869$ & $0.3798$ & $\textbf{0.3608}$ & $0.3443$ & $0.3373$ \\
      & 
      & 10 &  $\textbf{0.2612}$ & $0.2604$ & $0.2577$ & $\textbf{0.2158}$ & $0.2140$ & $0.2095$ \\
      & \multirow{2}{*}{TDDS}
      & 25 & $\textbf{0.2632}$ & $0.2615$ & $0.2557$  & $\textbf{0.2646}$ & $0.2611$ & $0.2559$ \\
      & 
      & 10  & $\textbf{0.2372}$ & $0.2275$ & $0.2219$ & $\textbf{0.2297}$ & $0.2238$ & $0.2184$ \\
    \cmidrule(lr){1-9}

    \multirow{4}{*}{Synthetic}
      & \multirow{3}{*}{DU}
      & 30 &  $0.5593$ & $0.5606$ & $\textbf{0.5634}$ & $0.7300$ & $\textbf{0.7344}$ & $0.7281$ \\
      & 
      & 20 &  $\textbf{0.4910}$ & $0.4886$ & $0.4829$ & $\textbf{0.7009}$ & $0.6983$ & $0.6942$ \\
      & 
      & 10 &  $\textbf{0.3396}$ & $0.3320$ & $0.3282$ & $\textbf{0.5593}$ & $0.5513$ & $0.5472$ \\
      & \multirow{2}{*}{TDDS}
      & 20 &  $\textbf{0.4236}$ & $0.4193$ & $0.4155$ & $\textbf{0.6713}$ & $0.6711$ & $0.6673$ \\
      & 
      & 10 & $\textbf{0.3849}$ & $0.3801$ & $0.3763$ & $\textbf{0.5722}$  & $0.5715$ & $0.5639$ \\
      \cmidrule(lr){1-9}

    \multirow{4}{*}{CIFAR-10}
      & \multirow{2}{*}{DU} 
      & 40 &  $0.6163$ & $\textbf{0.6318}$ & $0.6197$ & $0.6533$ & $\textbf{0.6608}$ & $0.6582$ \\
      &
      & 20 &  $\textbf{0.4080}$ & $0.2853$ & $0.2574$ & $\textbf{0.3915}$ & $0.3009$ & $0.2619$ \\
      & \multirow{2}{*}{TDDS} 
      & 40 &  $\textbf{0.3891}$ & $0.3217$ & $0.3007$ & $\textbf{0.5656}$ & $0.4213$ & $0.4076$ \\
      &
      & 20 &  $\textbf{0.1763}$ & $0.1630$ & $0.1571$ & $\textbf{0.2008}$ & $0.1120$ & $0.1095$ \\
    \bottomrule
  \end{tabular}
\end{table}

\section{Additional Experiments}

\subsection{Relationship between Correlation and Downstream Task Accuracy}
We investigate whether higher Pearson ($\rho$) or Spearman rank ($r_s$) correlations between extrapolated and ground truth scores are indicative of improved pruning performance, particularly in regimes where ground truth pruning outperforms random pruning. For the highest pruning rate at which ground truth pruning yields superior accuracy to random pruning, we compare the downstream accuracies obtained by retaining top-scoring samples according to various scoring methods (extrapolated, standard, and random). Note that random pruning corresponds to $0$ correlation, while ground truth scores correspond to perfect correlation ($\rho=1$, $r_s=1$). Our findings indicate that increased correlation between extrapolated and ground truth scores leads to downstream accuracies that closely match those of ground truth-based pruning (see \Cref{fig:CorrToAcc}). We further quantify the relationship between correlation metrics and downstream accuracy ($A$) after pruning by computing the Pearson and Spearman correlations between these variables. Results are summarized in \Cref{tab:corr_between_corr}.

\begin{table}[t]
\centering
\caption{Pearson ($\rho$) and Spearman ($r_s$) correlations of scores correlation between extrapolated and original scores, and post-pruning accuracy ($A$)}
\label{tab:corr_between_corr}
\begin{tabular}{c c c c c c c}
\toprule
Dataset           & Prune \% & Method & $\rho(\rho, A)$ & $\rho(r_s, A)$ & $r_s(\rho, A)$ & $r_s(r_s, A)$ \\
\midrule
\multirow{2}{*}{Places365} 
                  & 50       & DU     & 0.977           & 0.975          & 0.771          & 0.771         \\
                  & 95       & TDDS   & 0.914           & 0.914          & 1.000          & 1.000         \\
\cmidrule(lr){1-7}
ImageNet          & 50       & DU     & 0.766           & 0.779          & 0.771          & 0.771         \\
\cmidrule(lr){1-7}
\multirow{2}{*}{Synthetic CIFAR} 
                  & \multirow{2}{*}{95} 
                             & DU     & 0.995           & 0.944          & 0.943          & 0.943         \\
                  &          & TDDS   & 0.901           & 0.940          & 1.000          & 1.000         \\
\bottomrule
\end{tabular}
\end{table}

\subsection{Pruning Performance with Smaller Subset Sizes}
\begin{figure}[bt]
\vskip -0.3cm
\centering
\subfloat[Places 365\label{fig:appendixpruningPlaces}]{%
   \includegraphics[width=0.49\textwidth]{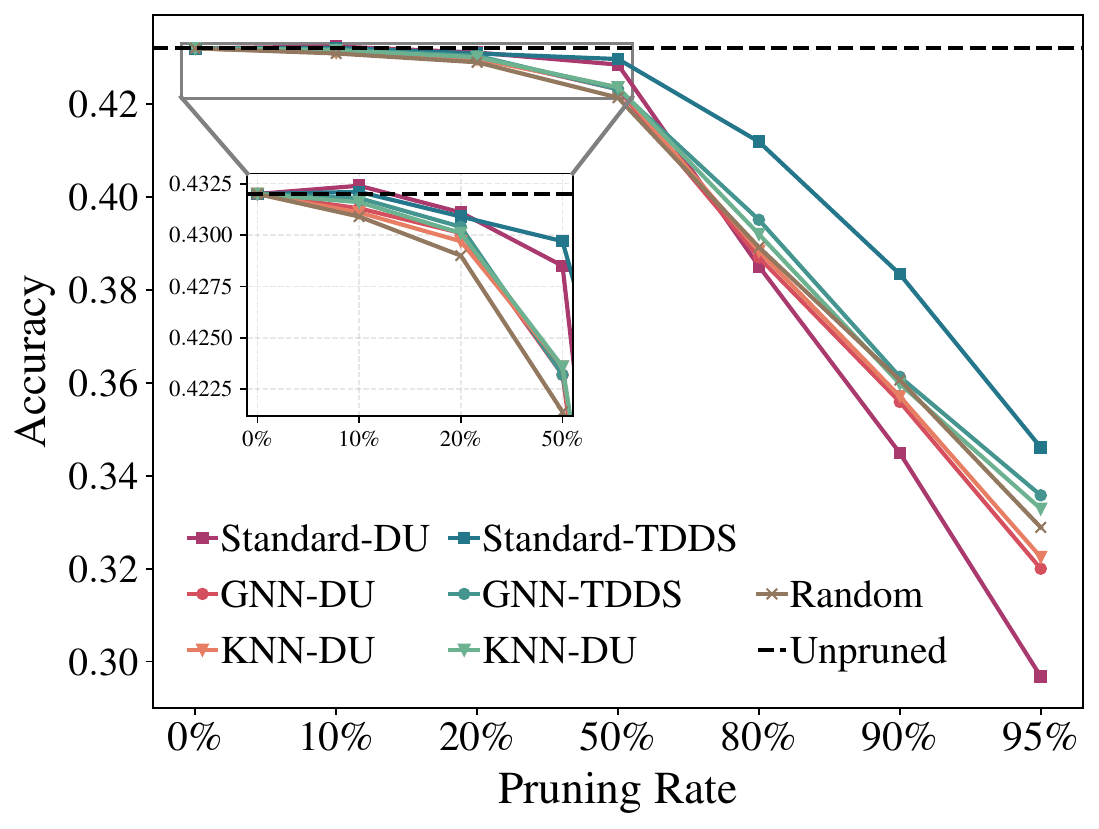}
}%
\subfloat[ImageNet\label{fig:appendixpruningImageNet}]{%
 \includegraphics[width=0.483\textwidth]{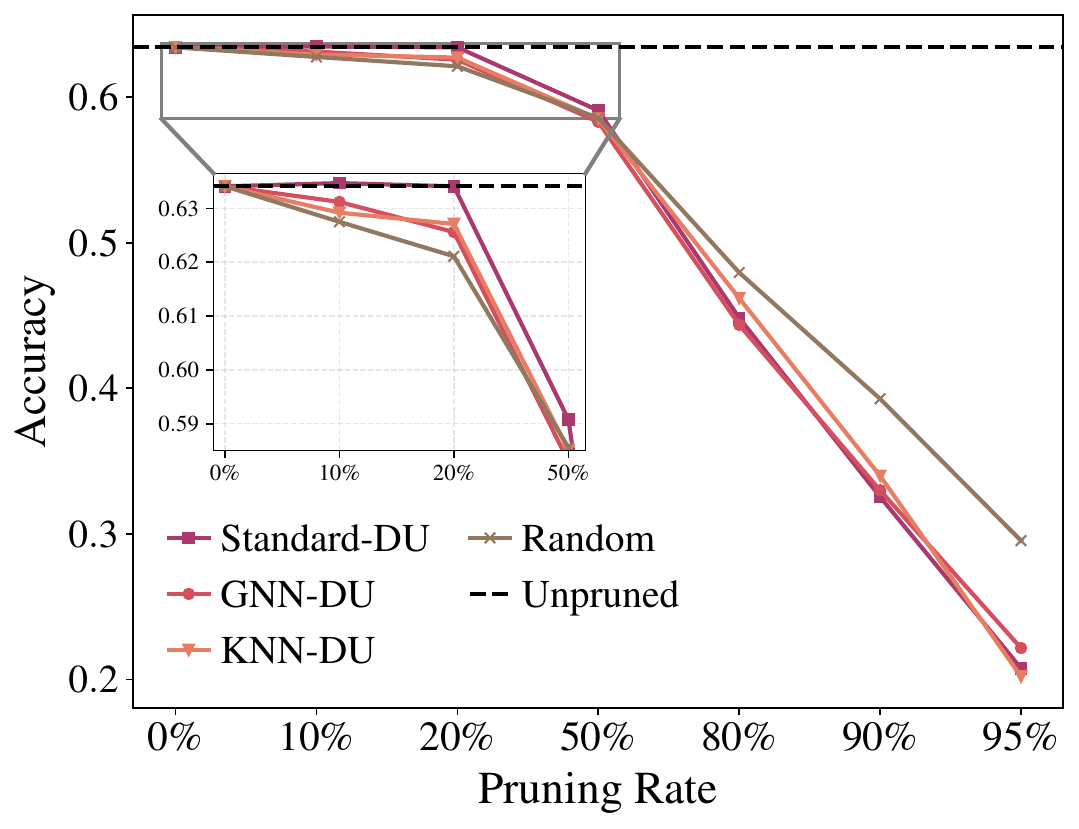}
}
\vskip -0.3cm

\subfloat[Synthic CIFAR-100\label{fig:appendixpruningCifar100}]{%
   \includegraphics[width=0.49\textwidth]{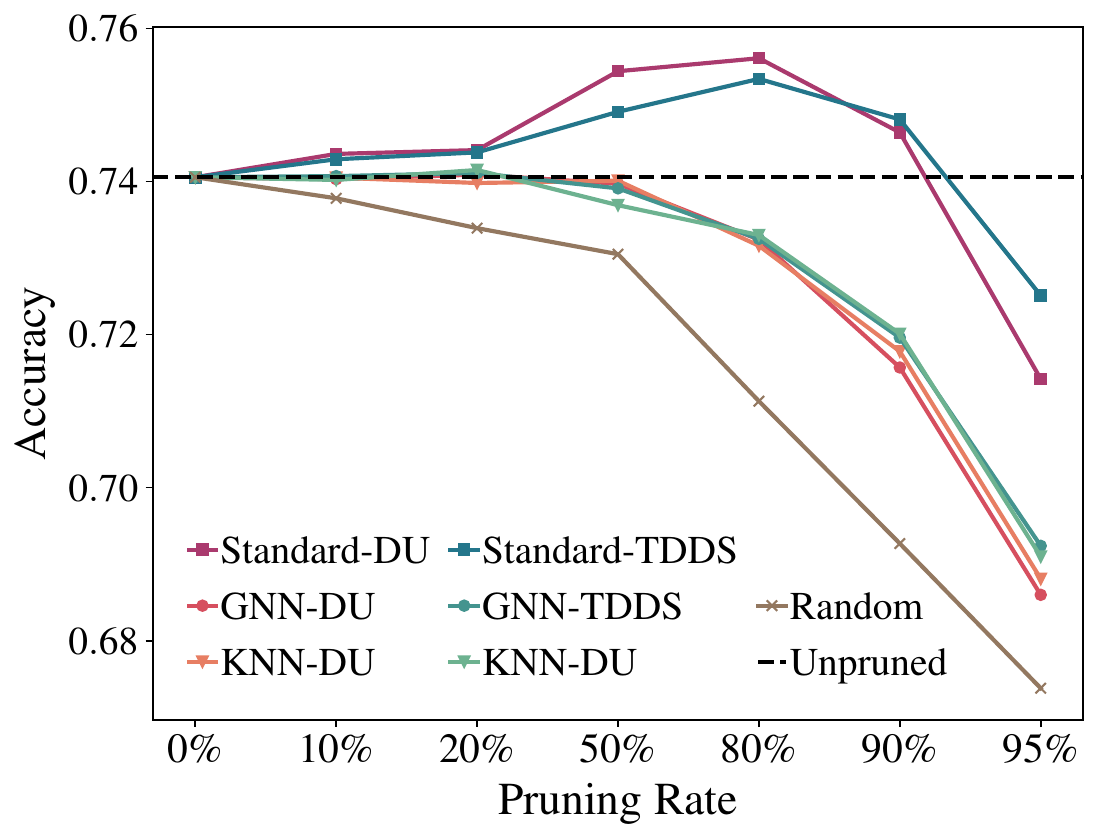}
}%
\subfloat[CIFAR-10\label{fig:appendixpruningCifar10}]{%
 \includegraphics[width=0.483\textwidth]{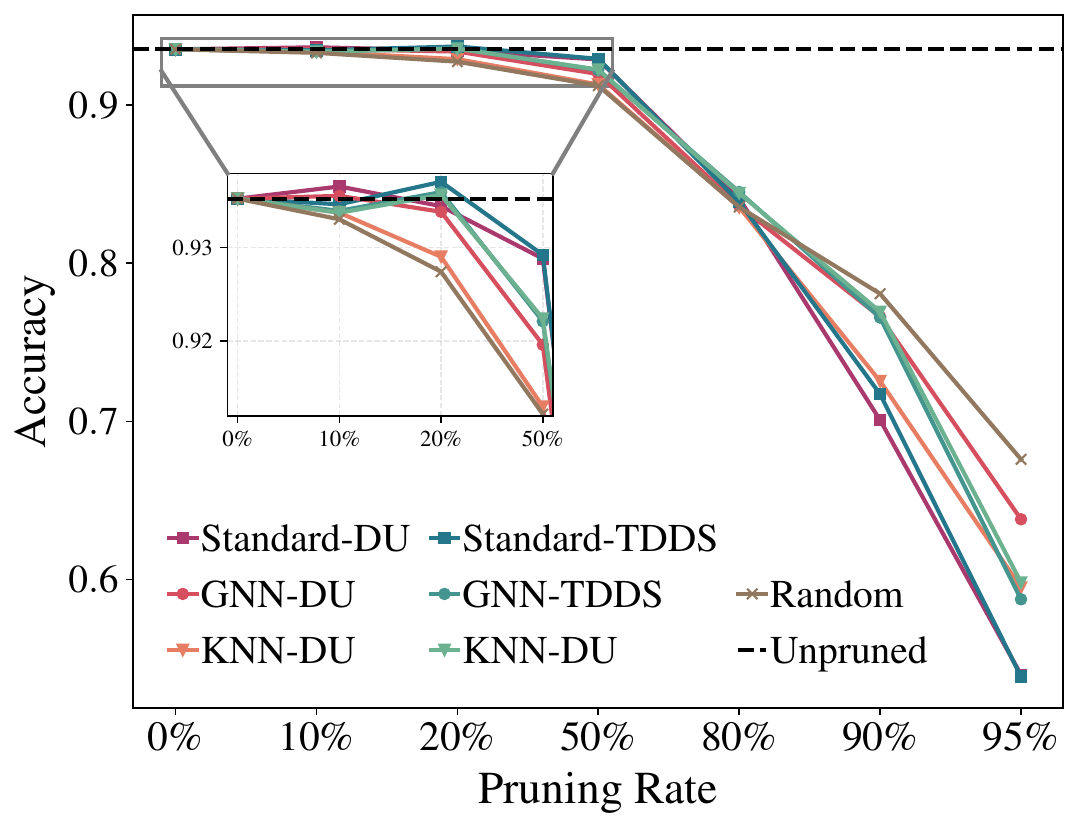}
}%
\vskip -0.1cm
\caption{Pruning performance of Standard approaches and their extrapolated counterparts, which started with score computation on $10\%$ subset ($20\%$ for CIFAR-10), and then extrapolated with KNN, and GNN. }
\label{fig:appendix_pruning}
\vskip -0.5cm
\end{figure}

In \Cref{fig:secondExperiment} we evaluated the pruning performance of Random Pruning, Standard pruning (DU, and TDDS), and extrapolation-based pruning with initial score computation on a subset of size $m = 20\%$ ($25\%$ for Places-365, $40\%$ for CIFAR-10). Here we provide additional results examining the impact of reducing the initial subset size, specifically considering $m=10\%$ ($20\%$ for CIFAR-10). The results are presented in \Cref{fig:appendix_pruning}.

We observe that, consistent with the initial larger subset size (\Cref{fig:secondExperiment}), pruning with extrapolated scores, even with a smaller initial subset size, outperforms the random baseline whenever the respective standard scores do. However, as the initial subset size $m$ decreases, the effectiveness of extrapolated scores diminishes. They also have smaller Pearsons correlation and Spearman rank with the standard score, as demonstrated in  \Cref{tab:correlation-results,tab:all_correlation_gnn,tab:all_correlation_knn}.

\newpage

\clearpage
\bibliography{main}

\end{document}